\pdfoutput=1
\documentclass[10pt,twocolumn,letterpaper]{article}

\usepackage{cvpr}              % To produce the CAMERA-READY version
\usepackage[dvipsnames]{xcolor}

\definecolor{cvprblue}{rgb}{0.21,0.49,0.74}

\usepackage[accsupp]{axessibility}

\usepackage{color,xcolor}
\usepackage{colortbl}
\usepackage{amssymb}
\usepackage{pifont}

\usepackage{tabu}
\usepackage{multirow}
\usepackage{amsmath,bm}

\usepackage{wrapfig}
\usepackage{adjustbox}
\usepackage[pagebackref,breaklinks,colorlinks,citecolor=cvprblue]{hyperref}

\newcommand{\first}[1]{\textcolor{red}{#1}}
\newcommand{\second}[1]{\textcolor{blue}{#1}}

\definecolor{gray}{rgb}{0.85, 0.85, 0.85}

\DeclareMathOperator{\jpeg}{JPEG}

\DeclareMathOperator{\concat}{Concat}
\DeclareMathOperator{\mlp}{MLP}

\DeclareMathOperator{\ode}{ODE}

\def\paperID{3731} % *** Enter the Paper ID here
\def\confName{CVPR}
\def\confYear{2024}

\title{FlowIE: Efficient Image Enhancement via Rectified Flow}

\def\spaces{~~~~~~}
\author{Yixuan Zhu\textsuperscript{1}\thanks{Equal contribution. ~~\textsuperscript{\dag}Corresponding author.}\spaces{}Wenliang Zhao\textsuperscript{1}\footnotemark[1]\spaces{}Ao Li\textsuperscript{2}\spaces{}Yansong Tang\textsuperscript{2}\spaces{}Jie Zhou\textsuperscript{1}\spaces{}Jiwen Lu$^{1,\dagger}$\\\\
\textsuperscript{1}Department of Automation, Tsinghua University \\%~~
\textsuperscript{2}Tsinghua Shenzhen International Graduate School, Tsinghua University
}

\begin{document}
\maketitle
\begin{abstract}
Image enhancement holds extensive applications in real-world scenarios due to complex environments and limitations of imaging devices. Conventional methods are often constrained by their tailored models, resulting in diminished robustness when confronted with challenging degradation conditions. In response, we propose FlowIE, a simple yet highly effective flow-based image enhancement framework that estimates straight-line paths from an elementary distribution to high-quality images. Unlike previous diffusion-based methods that suffer from long-time inference, FlowIE constructs a linear many-to-one transport mapping via conditioned rectified flow. The rectification straightens the trajectories of probability transfer, accelerating inference by an order of magnitude. This design enables our FlowIE to fully exploit rich knowledge in the pre-trained diffusion model, rendering it well-suited for various real-world applications. Moreover, we devise a faster inference algorithm, inspired by Lagrange's Mean Value Theorem, harnessing midpoint tangent direction to optimize path estimation, ultimately yielding visually superior results. Thanks to these designs, our FlowIE adeptly manages a diverse range of enhancement tasks within a concise sequence of fewer than 5 steps. Our contributions are rigorously validated through comprehensive experiments on synthetic and real-world datasets, unveiling the compelling efficacy and efficiency of our proposed FlowIE. Code is available at \url{https://github.com/EternalEvan/FlowIE}.
\end{abstract}    
\section{Introduction}
\label{sec:intro}

\begin{figure}[t]
    \centering
    \includegraphics[width=0.99\linewidth]{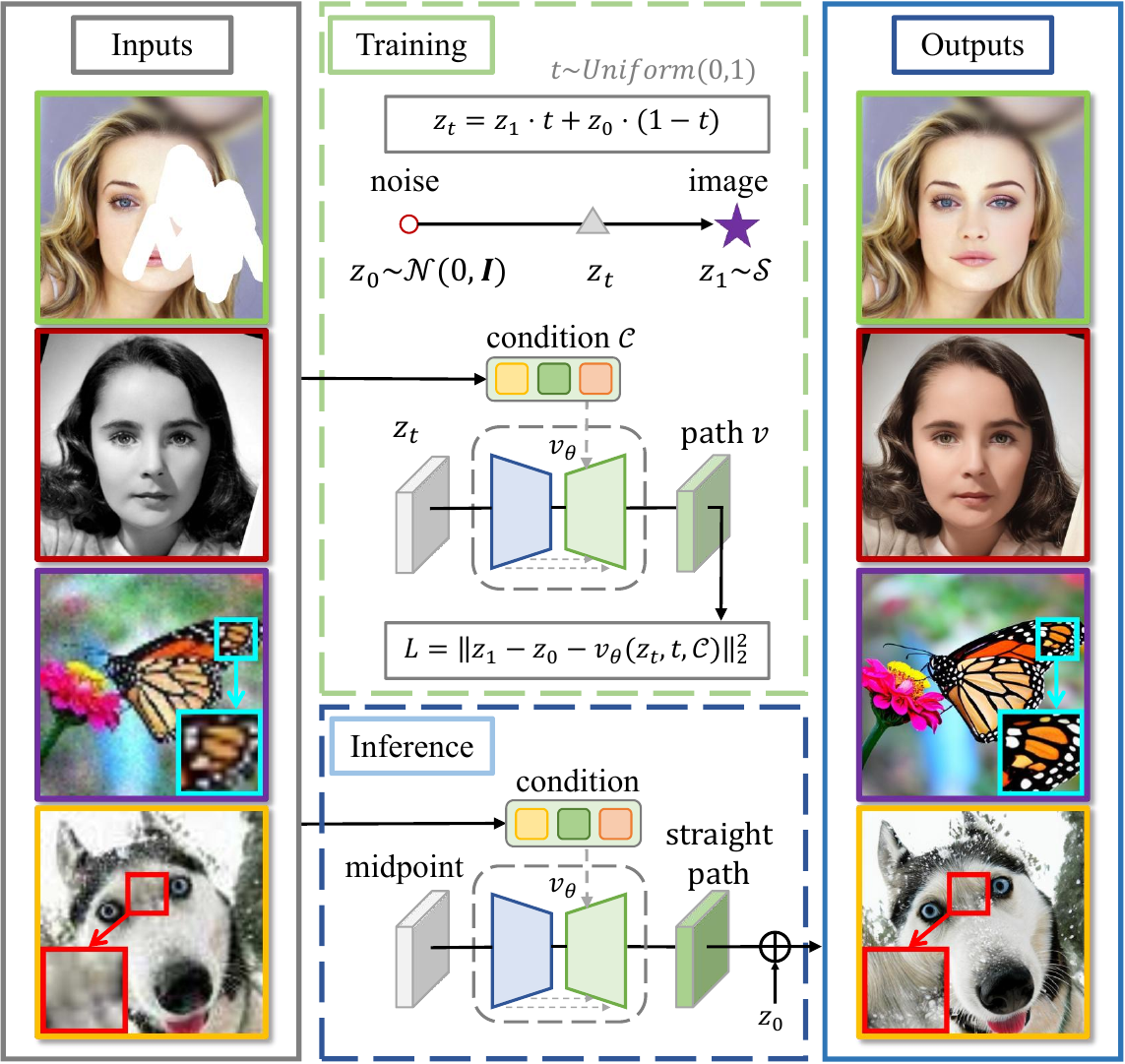}
    \caption{\textbf{The diagram of the proposed \textit{FlowIE}.} FlowIE leverages rectified flow to unleash the rich knowledge in the trained diffusion model and build straight-line paths between element distribution and clean images. The framework consistently achieves visually stunning results in a minimal number of steps and seamlessly generalizes to various image enhancement tasks, \textit{e.g.}, face inpainting, color enhancement and blind image super-resolution. }
    \label{fig:teaser}
    \vspace{-10pt}
\end{figure}
%-------------------------------------------------------------------------

The goal of image enhancement is to improve the visual quality of images afflicted by a wide range of factors, covering tasks like denoising, deblurring, super-resolution and inpainting. This field has garnered substantial attention for its vast utility in image restoration, camera designing, film-making and other domains. Recent years have witnessed notable advancements in image enhancement and a series of methods~\cite{dong2014learning,zhang2017beyond,liang2021swinir,chen2021pre,wang2022uformer,zamir2022restormer,chen2023activating}, based on deep learning has been introduced to produce high-quality outcomes. They exhibit commendable performance when confronted with specific and well-defined degradations. However, their efficacy becomes circumscribed when extended to intricate and unpredictable challenges posed by complex real-world scenarios. In practice terms, we aim to design a robust and efficient framework that excels in image enhancement, proficiently restoring general images affected by a diverse spectrum of real-world degradations.% This pursuit aligns more closely with the exigencies of real-world applications.

The challenge of image enhancement is fundamentally ill-posed, given the absence of explicit constraints governing the restoration process, thus permitting various plausible high-quality (HQ) results from the low-quality (LQ) inputs. To address this intricate problem, researchers explore approaches based on deep learning models that offer strong priors to guide the enhancement. We roughly categorize them into predictive~\cite{huang2020unfolding,gu2019blind,zhang2018learning}, GAN-based~\cite{wang2021unsupervised,yuan2018unsupervised,fritsche2019frequency,zhang2021designing,wang2021real} and diffusion-based~\cite{kawar2022denoising,wang2022zero,fei2023generative,lin2023diffbir} methods. Predictive methods seek to explicitly model the blur kernel from LQ images and restore HQ images with these predicted parameters. However, their adaptability to the complexity of real-world conditions remains limited due to the simple degradation setting and the vulnerable estimated results. To improve the enhancement quality, some approaches employ the Generative Adversarial Network (GAN)~\cite{goodfellow2014generative} to implicitly learn the data distribution and degradation model. GAN-based methods like~\cite{zhang2021designing} and~\cite{wang2021real} achieve considerable results with the image priors from GANs and high-order degradation models. Nevertheless, the tuning of GAN-based methods poses a persistent challenge attributed to their complex losses and hyper-parameters. More recently, Diffusion Models (DMs)~\cite{ho2020denoising,rombach2022high} have demonstrated remarkable capabilities in synthesizing visually compelling images. Following this line, some methods leverage the strong generative prior from the pre-trained diffusion model to attain high-quality restorations. For example,~\cite{kawar2022denoising},~\cite{wang2022zero} and~\cite{fei2023generative} devise zero-shot techniques, which involve the direct utilization of diffusion model weights without training. Other methods like~\cite{lin2023diffbir,saharia2022image} fine-tune the diffusion model to better suit enhancement tasks. Though diffusion-based methods yield impressive outcomes, they are hampered by high computational demands and protracted inference times, rendering them less practical for industrial applications. 
\begin{figure}[t]
    \centering
    \includegraphics[width=0.99\linewidth]{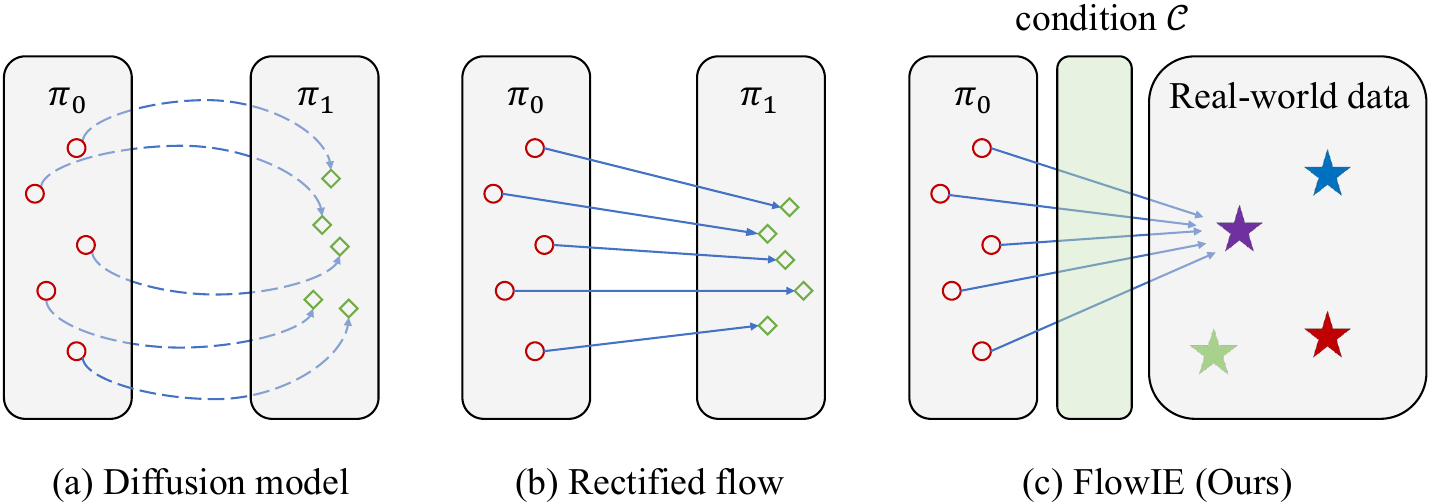}
    \caption{
    As shown in (a), diffusion models~\cite{ho2020denoising,rombach2022high} solve ODEs in curve trajectories. Differently, rectified flow~\cite{liu2022flow}, illustrated in (b), bridges one-to-one straight paths between two distributions, thereby reducing the inference steps. (c) Our proposed FlowIE applies the flow-based framework to real-world data and discards the massive data preparation process. We construct a many-to-one mapping that predicts straight paths to clean images from arbitrary noise in an elementary distribution with proper guidance.}
    \label{fig:cmp_method }
    \vspace{-15pt}
\end{figure}

    %\textbf{Comparison of the proposed \textit{FlowIE}.} Unlike diffusion models~\cite{ho2020denoising,rombach2022high}, which solve ODEs in curve trajectories, rectified flow~\cite{liu2022flow} bridges one-to-one straight paths between two distributions, thereby reducing the inference steps. FlowIE applies the flow-based framework to real-world data and discards the massive data pair preparation process. We construct a many-to-one mapping that predicts straight paths to clean images from arbitrary noise in an element distribution with the proper guidance.

To overcome the aforementioned challenges, we propose a simple yet potent framework named FlowIE for diverse real-world image enhancement tasks. FlowIE dramatically reduces the inference time by a magnitude of tenfold compared to diffusion-based methods while upholding the exceptional quality of the enhancements, as showcased in Figure~\ref{fig:teaser}. Our primary objection is to harness the generative priors of pre-trained diffusion for restoring images beset by general degradation. Diverging from existing diffusion-based methods, we abandon the extensive denoising steps in diffusion sampling via conditioned rectified flow. This approach straightens the trajectories of probability transfer during diffusion sampling, offering a swift and unified solution for the distribution transport in diffusion models. Once the straight path from an elementary distribution to the real-world HQ image is accurately estimated, we can yield a computationally efficient model since it is the shortest path between two points. However, rectified flow cannot directly adapt to enhancement tasks, given its one-to-one noise-image mapping paradigm and reliance on training with synthetic images that differ significantly from real-world data. To address these limitations, we employ rectified flow to predict paths from any noise to one real-world image, thus constructing a many-to-one transport mapping, as shown in Figure~\ref{fig:cmp_method }. This relaxation of rectified flow allows us to avoid the expansive data pair preparation process and effectively unleashes the generative potential inherent in the pre-trained diffusion model for image enhancement tasks. After learning from real-world data, our framework can determine a nearly straight path toward the target result. To further refine the prediction accuracy, we devise a mean value sampling inspired by Lagrange's Mean Value Theorem to estimate the path with higher precision from a midpoint along the transport path. %We illustrate that our correction strategy not only reduces the inference steps but also significantly improves the overall result quality.

We mainly evaluate our method on two representative image enhancement tasks covering: 1) blind face restoration (BFR) and 2) blind image super-resolution (BSR). We show that our FlowIE can play as an effective enhancer for degraded images, effectively catering to a broad spectrum of tasks. Our model attains 19.81 FID and 0.69 IDS on synthetic CelebA-Test~\cite{liu2015deep}, establishing new state-of-the-art on these two benchmarks. On real-world LFW-Test~\cite{wang2021towards} and WIDER-Test datasets~\cite{zhou2022towards}, our model achieves 38.66 and 32.41 FID, respectively, exhibiting high restoration quality in the real-world condition. After tuning on ImageNet~\cite{deng2009imagenet}, we obtain 0.5953 MANIQA on RealSRSet~\cite{cai2019toward} and 0.6087 on our Collect-100, highlighting our effectiveness in general image restoration. Except for higher metrics compared with other diffusion-based methods, we showcase an almost 10 times faster inference speed thanks to rectified flow. To explore the potential of FlowIE on further tasks, we extend the application of FlowIE to face color enhancement and inpainting with only 5K steps of fine-tuning. FlowIE consistently delivers visually appealing and plausible enhancements, underscoring its robust generalization capability.
%We expect our study to provide a new perspective for flow-based applications on low-level vision tasks and inspire further research on both faster diffusion sampling and image enhancement.

\section{Related Works}
\label{sec:formatting}

\noindent\textbf{Predictive Methods.} Image enhancement consists of various manipulations and refinements, including denoising, super-resolution (SR), inpainting, \textit{etc}. Some works utilize predictive models to address these tasks. Notably, convolution-based methods~\cite{dong2014learning,huang2020unfolding,zhang2018learning,dong2015image} adopt explicit approaches for SR task by estimating the blur kernels and restoring HQ images with the predicted kernels. On the other hand, With the advent of vision transformers~\cite{vit,swin}, some methods~\cite{chen2021pre,liang2021swinir} propose frameworks incorporating attention-based architectures, yielding high-quality results on SR, denoising, and deraining tasks. Although predictive approaches pave the way for various enhancement tasks, they still struggle to handle complex real-world conditions due to their simple degradation settings during training.

\noindent\textbf{GAN-based Methods.} In addition to predictive methods, another line of work explores employing generative models like GAN~\cite{goodfellow2014generative} to provide embedded image priors. GAN-based methods~\cite{wang2021unsupervised,yuan2018unsupervised,fritsche2019frequency,zhang2021designing,wang2021real} learn how to process images in the latent space, showcasing notable achievements in tasks such as BSR. Moreover, works like~\cite{wang2021towards,yang2021gan,zhou2022towards} leverage GAN priors for the BFR task and yield satisfying outcomes. However, GAN-based methods exhibit some drawbacks like the potential for unstable results and the necessity for meticulous hyper-parameter tuning. Furthermore, their architectures are often tailored to specific tasks, limiting their adaptability across diverse applications.

\noindent\textbf{Diffusion-based Methods.}
Diffusion models~\cite{ho2020denoising,rombach2022high} are well-known for their powerful image synthesis capability and robust training procedure. To harness the image priors of the pre-trained diffusion model, methods such as~\cite{kawar2022denoising,wang2022zero,fei2023generative} propose training-free approaches to enhance image quality in a zero-shot manner, showcasing the adaptability of diffusion models across various tasks. In a parallel line of research, supervised approaches like~\cite{saharia2022image,lin2023diffbir,wang2023exposurediffusion} pave the way to fine-tune the diffusion model, improving its generative potential. Despite achieving visually appealing results, diffusion-based methods suffer from long-time sampling due to repeated model inference. To mitigate the time-consuming problem and fully exploit the generative priors within the pre-trained diffusion model, we devise a novel flow-based framework designed for diverse image enhancement tasks. Our framework utilizes the rich knowledge from the pre-trained diffusion model and accelerates the inference via a flow-based approach, which straightens the transport trajectories from an elementary distribution to real-world data and thereby realizes efficient inference.
 %constructs a many-to-one mapping toward real-world images, 
%Recent advancements in rectified flow~\cite{liu2022flow,liu2022rectified} introduce an efficient strategy to solve the transport mapping problems within diffusion sampling by building straight line paths between an element distribution and the synthetic images, facilitating a much faster inference. However, existing flow-based generative methods encounter challenges in building extensive training data pairs through repeated diffusion sampling. Moreover, these models struggle to directly adapt to enhancement tasks, given their one-to-one noise-image mapping paradigm and reliance on training with synthetic images that diverge significantly from real-world data. To employ flow-based methods for efficient image enhancement, we must reconstruct the transport mapping and control the generative priors with proper conditions. % 
%-------------------------------------------------------------------------
%\subsection{Latent Diffusion Models}

%\subsection{Rectified Flow}
\begin{figure*}[htp]
    \centering
    \includegraphics[width=0.98\linewidth]{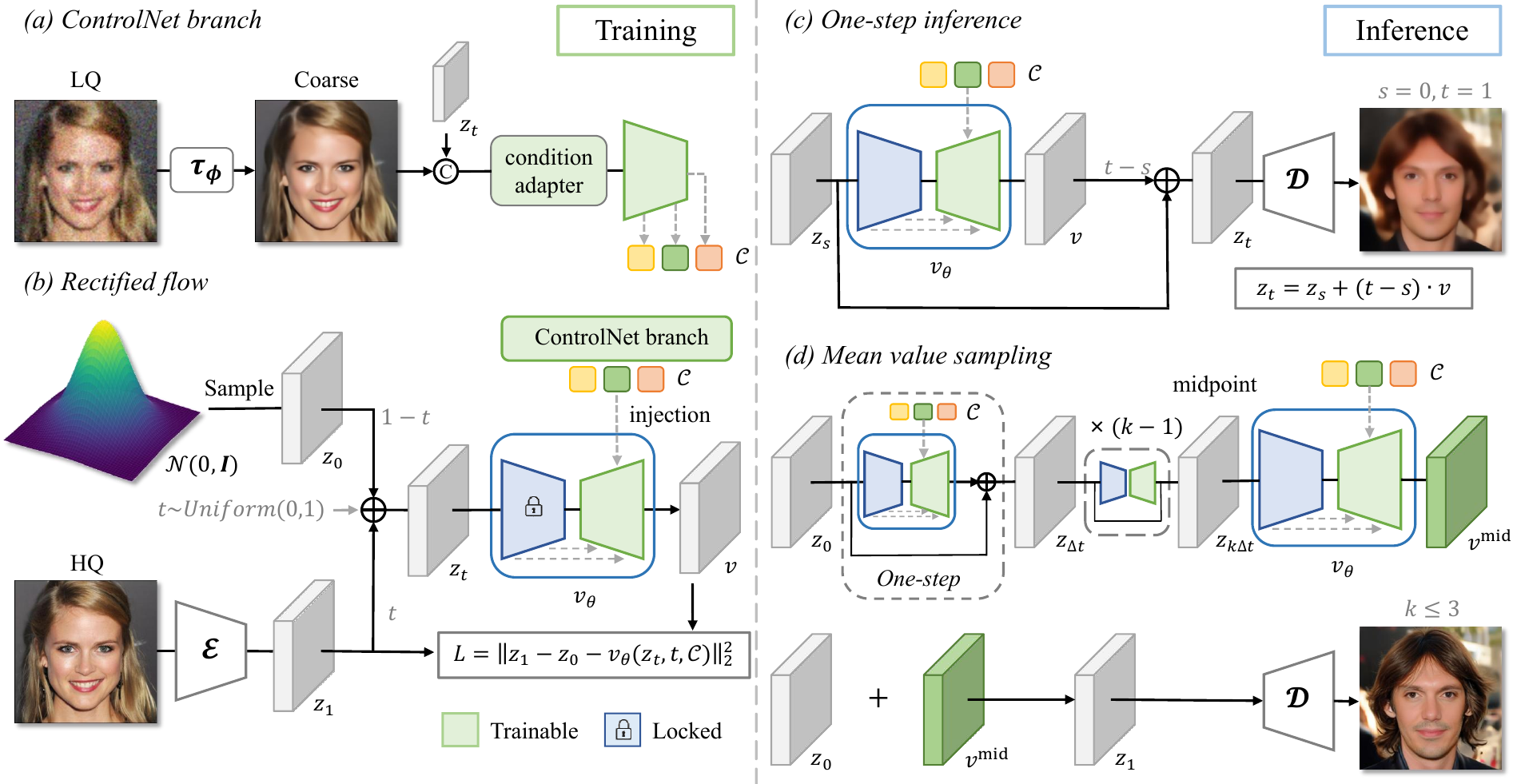}
    \caption{\textbf{The overall framework of \textit{FlowIE}.} FlowIE is a flow-based framework for image enhancement tasks. During training (left), we optimize the rectified flow ${\bm v}_\theta$ to bridge straight line paths ${\bm v}$ from an elementary distribution to clean images with proper guidance. We also developed the mean value sampling to improve the path estimation. During inference (right), we utilize the conditions from LQ to predict a linear direction toward the clean images on the midpoint of the transport curve, yielding high-quality and visually appealing results. }%To fully exploit the rich knowledge from image generation pre-training, we employ rectified flow with proper guidance to bridge a straight line path between element distribution and clean images. Our framework can achieve various}
    \label{fig:pipeline}
    \vspace{-10pt}
\end{figure*}

%-------------------------------------------------------------------------

%-------------------------------------------------------------------------

%-------------------------------------------------------------------------

%Our key idea is to fully exploit the generative prior of the trained diffusion model for image enhancement tasks and accelerate the inference process via a flow-guided straight path toward clean images.

\section{Method}
In this section, we present FlowIE, a simple flow-based framework that fully exploits the generative diffusion prior for efficient image enhancement. We will start by providing a brief background on rectified flow and then delve into the key designs of FlowIE. This includes the construction of a flow-based enhancement model, the design of appropriate conditions as guidance, quality improvements through mean value sampling, and detailed implementations. The overall pipeline of FlowIE is depicted in Figure~\ref{fig:pipeline}.
\subsection{Preliminaries: Rectified Flow}
We commence by briefly introducing rectified flow~\cite{liu2022flow,liu2022rectified}. Rectified flow is a series of methods for solving the transport mapping problem: given observations of two distributions $X_0\sim\pi_0$, $X_1\sim\pi_1$ on $\mathbb{R}^d$, find a transport map $T:\mathbb{R}^d\rightarrow\mathbb{R}^d$, such that $T(X_0)\sim\pi_1$ when $X_0\sim\pi_0$. 
%Previous methods utilize generative models like GAN and VAE to parameterize $T$ using neural networks, yielding great progress in image editing and synthesis. 
Diffusion models represent transport mapping problems as a continuous time process governed by stochastic differential equations (SDEs) and leverage a neural network to simulate the drift force of the processes. The learned SDEs can be transformed into marginal-preserving
probability flow ordinary differential equations (ODEs)~\cite{song2020denoising,song2020score} to facilitate faster inference. However, diffusion models still suffer from long-time sampling due to repeated network inference to solve the ODEs/ SDEs, compared to one-step models like GANs. To address this problem, rectified flow introduces an ODE model that transfers $\pi_0$ to $\pi_1$ via a straight line path, theoretically the shortest route between two points:
\begin{equation}
    {\rm d}X_t={\bm v}(X_t,t){\rm d}t,
\end{equation}
where $\bm v$ represents the velocity guiding the flow to follow the direction of $(X_1-X_0)$ and $t\in[0,1]$ denotes the time of the process. To estimate ${\bm v}$, rectified flow solves a simple least squares regression problem that fits $\bm v$ to $(X_1-X_0)$.

In practice, rectified flow leverages a network ${\bm v}_\theta$ to predict the velocity (path direction), and draws data pairs $\mathcal{X}=\{(X_0, X_1)|X_1=\ode(X_0)\}$, where $\ode$ denotes a trained diffusion model, to minimize the loss function $L$:
\begin{equation}
    \begin{split}
     L=\int_0^1\mathbb{E}_{X_0,X_1}&\left[\Vert (X_1-X_0)-{\bm v}_\theta(X_t,t)\Vert^2\right]{\rm d}t,
     \\
    X_t&=tX_1+(1-t)X_0.   
    \end{split}
    \label{flow_eq}
\end{equation}
With the optimized ${\bm v}_\theta$ as a path predictor, rectified flow bridges the gap between two distributions with almost straight paths. Using the forward Euler method, rectified flow induced from the training data can produce high-quality results with a small number of steps. 
%The reflow model (2-rectified flow) induced from 1-rectified flow exhibits even straighter trajectories and can yield results in only one step.
\subsection{Flow-based Image Enhancement}
The pre-trained diffusion model encompasses rich information about real-world data distribution and detailed image synthesis capacity. Our goal is to exploit the generative prior of a pre-trained diffusion model for image enhancement and mitigate the extensive computational cost of diffusion sampling. Our core idea involves adopting a rectified flow framework with proper guidance to tune the denoising U-Net ${\bm\epsilon}_\theta$~\cite{ronneberger2015u} in the text-to-image pre-trained diffusion model into an effective path predictor ${\bm v}_\theta$. This predictor enables the establishment of straight pathways from a simple elementary distribution to clean images, thereby facilitating the efficient utilization and acceleration of learned diffusion priors for image enhancement. In pursuit of this goal, we first define the image degradation models as ${\bm y}=\mathcal{D}_{h}(\bm x)$, where ${\bm x}$ and ${\bm y}$ are HQ and LQ images and $h\in\mathcal{H}$ denotes a specific enhancement task. Enhancing images from real-world degradation poses a significant challenge due to the inherent complexity of $\mathcal{D}_{ h}$ that is hard to formulate.  
%%%%%%%%%%%%%%%%%% 
To furnish precise and proper guidance for rectified flow within intricate scenarios, we employ a pre-trained initial-stage model ${\bm \tau}_\phi$ for coarse restoration. ${\bm \tau}_\phi$ is dedicated to blur reduction and contributes to the construction of the condition $\mathcal{C}$, which is pivotal for narrowing the direction spectrum and facilitating path prediction. Compared to the denoising U-Net of the diffusion model, ${\bm \tau}_\phi$ holds much fewer parameters and exerts minimal impact on inference speed.

In contrast to the image synthesis that rectified flow is typically tailored for, image enhancement tasks have a relatively deterministic target (HQ). Therefore, the one-to-one transport mapping inherent in rectified flow cannot directly apply to our work. Instead, we naturally consider a novel many-to-one mapping that every point in an elementary (Gaussian) distribution orients to a fixed HQ image in the real world. This method offers two advantages compared to~\cite{liu2022flow}: (1) it evades expensive data preparing that entails drawing massive data pairs by performing diffusion sampling repeatedly, and intuitively aligns with enhancement tasks with ground truth HQ data; (2) it stabilizes training and inference with theoretically infinite data pairs for learning and fully leverages the condition to control the rectified flow process. Motivated by these properties, we aim to design an effective control plan to centralize the flow direction, using the coarse result as the raw material. For this control mechanism, we employ a ControlNet branch, which consists of a condition adapter and an injection module, to introduce spatial guidance for the path predictor. Given a clean image ${\bm z}_1$ from real-world dataset $\mathcal{S}$ and a noise ${\bm z}_0$ sampled from the standard Gaussian distribution, we synthesize the LQ image ${\bm z}_{\rm LQ}$ using the degradation model and recover the coarse result with ${\bm \tau}_\phi$. To construct the condition $\mathcal{C}$ with the information of time $t$, we concatenate the coarse result with the noisy image ${\bm z}_t$ produced by the linear interpolation in Equation (\ref{flow_eq}). Then we employ the condition adapter implemented as a two-layer MLP to refine the image features and apply a zero convolution layer $\mathcal{F}$ with both weights and bias initialized to zeros to align the channel dimension. To sum up, the condition $\mathcal{C}$ is computed as:
\begin{equation}
    \begin{split}      
    \mathcal{C}&\leftarrow \concat({\bm \tau}_\phi({\bm z}_{\rm LQ}),{\bm z}_t)\\
    \mathcal{C}&\leftarrow \mathcal{C}+\gamma\mlp(\mathcal{C})\\
    \mathcal{C}&\leftarrow \mathcal{F}(\mathcal{C},t),
    %\vspace{-10 pt}
    \end{split}
    %\vspace{-10 pt}
\end{equation}
where $\gamma$ is a learnable scale factor initialized to be very small (\textit{e.g.}, 1e-4). Then we corporating the information in $\mathcal{C}$ to the denoising U-Net via the injection module. Given the time step $t$ and condition $\mathcal{C}$, our rectified flow optimize ${\bm v}_\theta$ to predict the straight line path by minimizing $L$:
\begin{equation}
    L=\mathbb{E}_{t,{\bm z}_1,{\bm z}_0\sim\mathcal{N}(0,{\bm I})}\left[\Vert{\bm z}_1-{\bm z}_0-{\bm v}_\theta({\bm z}_t,t,\mathcal{C})\Vert^2\right].
\end{equation}
The training procedure should be stable and quick benefiting from the inclusion of zero convolution layers, which prevent harmful noise for neural network layers and injects appropriate conditions into the rectified flow.

\subsection{Improve Quality via Mean Value Sampling}
The optimized rectified flow straightens the transport trajectories to nearly linear paths. Utilizing the forward Euler method, rectified flow can produce plausible results with a small number of Euler steps. However, a simple iterative forward method inevitably causes error accumulation, leading to global blur and unsatisfactory details. To tackle this problem, we devise the Mean Value Sampling based on Lagrange's Mean Value Theorem to improve the velocity estimation accuracy, yielding more visual-appealing results with better quality and details. Specifically, Lagrange's Mean Value Theorem states that for any two points on a curve, there exists a point on this curve such that the derivative of the curve at this point is equal to the scope of the straight line connecting these points. Since rectified flow acts as a path or derivative predictor on a differentiable transport curve, we naturally leverage it to find a midpoint on the curve that the velocity direction ${\bm v}^{\rm mid}$ of this point is parallel to the straight line bridging ${\bm z}_0$ and ${\bm z}_1$. 

For midpoint searching, we compute the path direction of a point set $\mathcal{P}=\{{\bm z}_0, {\bm z}_{\Delta t}, ..., {\bm z}_{1-\Delta t}\}$, covering uniform discrete timesteps along the curve with the step length $\Delta t=\frac{1}{N}$. We observe that there exists a midpoint ${\bm z}_{k\Delta t}$ in $\mathcal{P}$ that predicts the most accurate direction and yields the best result. We select the desirable $k\in\{0, 1,..., N-1\}$ with a few test data points for different tasks and we find that ${\bm z}_{k\Delta t}$ always produces reliable results in a specific task.
\subsection{Implementation}
We consider four image enhancement tasks with different degradation models $\mathcal{D}_h$ in this work. In detail, the degradation model for BFR and BSR can be generally approximated as ${\bm y}=\left[({\bm k}*{\bm x})\downarrow_r+{\bm n}\right]_{\jpeg}$, which consists of blur, noise, resize and JPEG compression. Since images usually suffer from more severe harassment in the real-world scene, we apply a high-order degradation model, repeating the above process multiple times. For the inpainting task, we need to recover the missing pixels in images. The corresponding degradation model is the dot-multiplication with a binary mask: ${\bm y}={\bm x}\odot{\bm m}$. For color enhancement, the degraded image experiences color shifts or only retains the grayscale channel. In our framework, we mainly manipulate images in a latent space constructed by a trained VQGAN, consisting of an encoder $\mathcal{E}$ and a decoder $\mathcal{D}$ and achieving the conversion between the pixel space and the latent space. We also create a trainable copy of the encoding blocks and the middle block in ${\bm v}_\theta$ as the injection module to handle the condition $\mathcal{C}$ and infuse it to ${\bm v}_\theta$. For optimal results, we empirically capture the midpoint with $N=5$ and $k=3$, thus the inference requires only $k+1=4$ steps.
\begin{table*}[!t]
  \centering  
  \caption{\textbf{Quantitative comparisons for BFR on the synthetic and real-world datasets.} \first{Red} and \second{blue} indicate the best and the second
best performance, respectively. We categorize the methods into conventional (up), diffusion-based (middle) and flow-based (bottom). Our FlowIE shows very competitive results compared with existing methods. We obtain remarkable image quality and identity consistency with the leading FID and IDS scores. Our framework also exhibits much faster inference than the diffusion-based method.}
\adjustbox{width=\linewidth}
  {\begin{tabu} to 1.2\linewidth {l*{9}{X[c]}}
    \toprule
    \multirow{3}{*}{Method}& \multicolumn{3}{c}{Wild Datasets} & \multicolumn{5}{c}{Synthetic Dataset} & \multirow{3}{*}{FPS$\uparrow$}\\
    & LFW & WIDER & CelebChild &\multicolumn{5}{c}{CelebA}    \\
    %\cline{3-10}
    \cmidrule(lr){2-4}\cmidrule(lr){5-9}
      & FID$\downarrow$ & FID$\downarrow$ & FID$\downarrow$&  PSNR$\uparrow$ & SSIM$\uparrow$ & LPIPS$\downarrow$ & FID$\downarrow$ & IDS $\uparrow$ \\
    \midrule
    GPEN~\cite{yang2021gan}  & $51.95$ & $46.41$ & $76.62$ & $21.3941 $ & $0.5745$ & $0.4685$ & $23.88$ & $0.49$ &7.278 \\
    GCFSR~\cite{he2022gcfsr}  & $52.18$ & $40.89$ & $76.32$ & $21.8789 $ & \second{$0.6070$} & $0.4579$ & $35.52$ & $0.45$ &9.243\\
    GFPGAN~\cite{wang2021towards}  & $52.11$ & $41.70$ & $80.69$ & $21.6953 $ & $0.6060$ & $0.4304$ & $21.69$ & $0.49$ &8.152  \\
    VQFR~\cite{gu2022vqfr}  & $49.92$ & $37.89$ & $74.75$ & $21.3012 $ & \first{$0.6125$} & \second{$0.4127$} & $20.47$ & $0.48$ &3.837 \\
    RestoreFormer~\cite{wang2022restoreformer} & $48.41$ & $49.82$ & \first{$71.09$} & $21.0029 $ & $0.5289$ & $0.4791$ & $43.76$ & $0.55$  &4.964 \\
    DMDNet~\cite{li2022learning} & $43.38$ & $40.53$ & $79.37$ &$21.6620 $ & $0.5997$ & $0.4825$ & $64.21$ & \second{$0.66$}  &3.454 \\
    CodeFormer~\cite{zhou2022towards}  & $52.34$ & $38.79$ & $79.58$ & \first{$22.1513$} & $0.5949$ & \first{$0.4057$} & $22.23$ & $0.48$ &5.188 \\
    \midrule
    DiffBIR~\cite{lin2023diffbir} & \second{$39.61$} & \second{$33.51$} & $77.74$ & $21.7512$ & $0.5968$ & $0.4575$ & \second{$20.19$} & $0.52$&0.285  \\
    \midrule
    FlowIE (Ours)  & \first{$38.66$} & \first{$32.41$} & \second{$74.25$} & \second{$21.9211$} & $0.6005$ & $0.4367$ & \first{$19.81$} & \first{$0.69$}  &2.846\\
    \bottomrule
  \end{tabu}}
  
  \label{tab:bfr_quant}
\end{table*}

\begin{figure*}[t]
    \centering
    \includegraphics[width=0.99\linewidth]{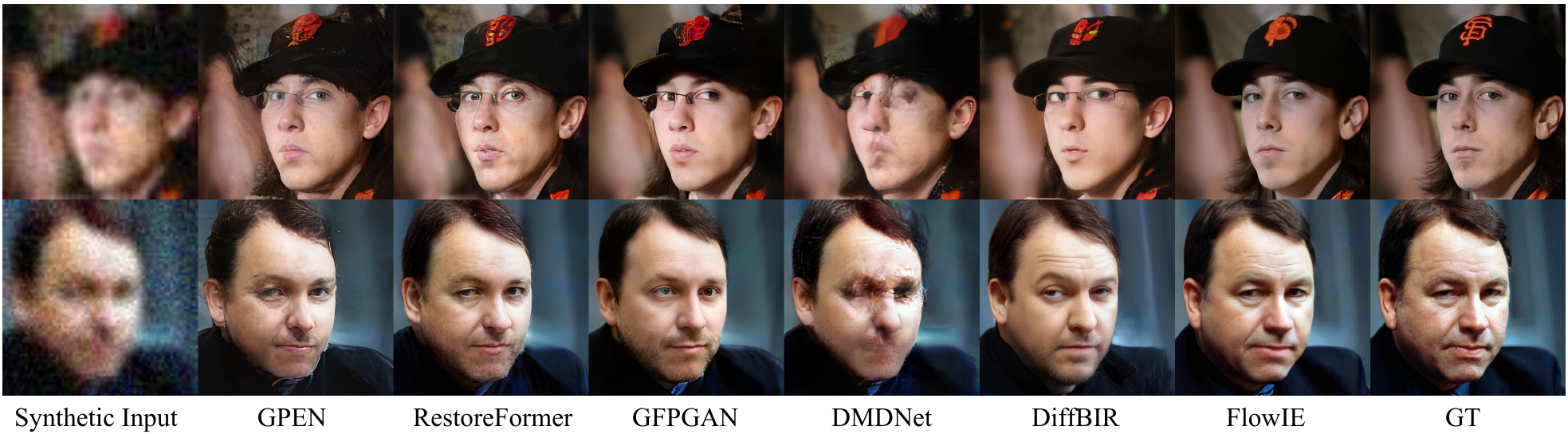}
    \vspace{-5pt}
    \caption{\textbf{Qualitative comparisons on CelebA-Test.} FlowIE generates plausible HQ results with enough details and high identity similarity even though input faces are severely degraded, while previous methods produce visible artifacts or inconsistent faces.}
    \label{fig:syn}
    \vspace{-10pt}
\end{figure*}
\begin{figure*}[t]

    \centering
    \includegraphics[width=0.99\linewidth]{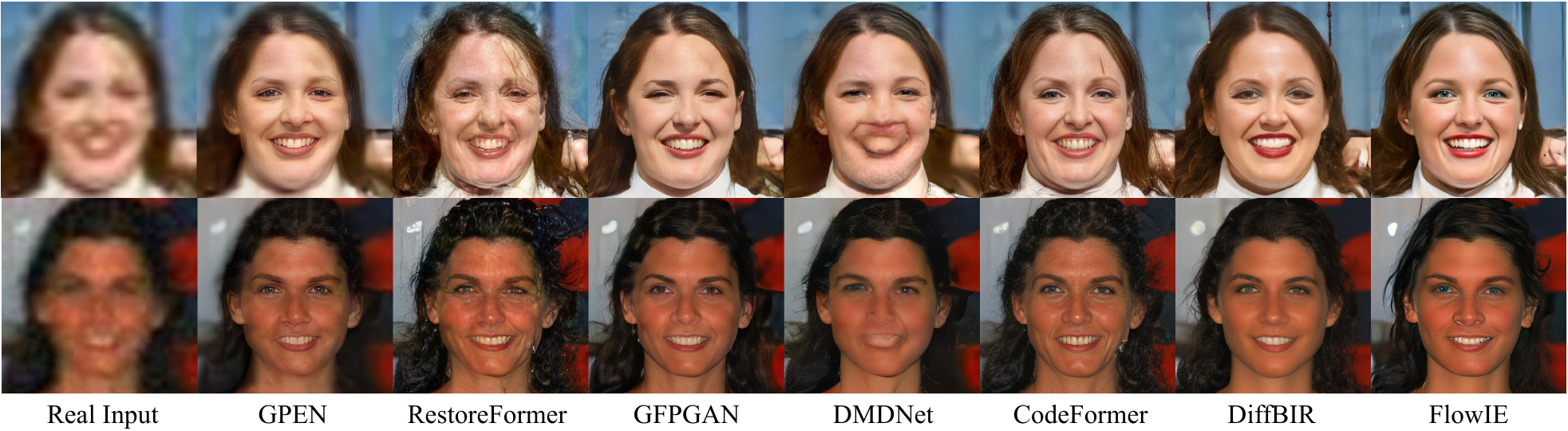}
    \vspace{-5pt}
    \caption{\textbf{Qualitative comparisons on real-world faces.} Our method performs plausible enhancement on real-world faces, producing high-fidelity and visually satisfactory faces. Compared to other methods, FlowIE enjoys robustness in front of challenging cases.}

    \label{fig:wild}
    \vspace{-5pt}
\end{figure*}
\begin{figure*}[t]
    \centering
    \includegraphics[width=0.99\linewidth]{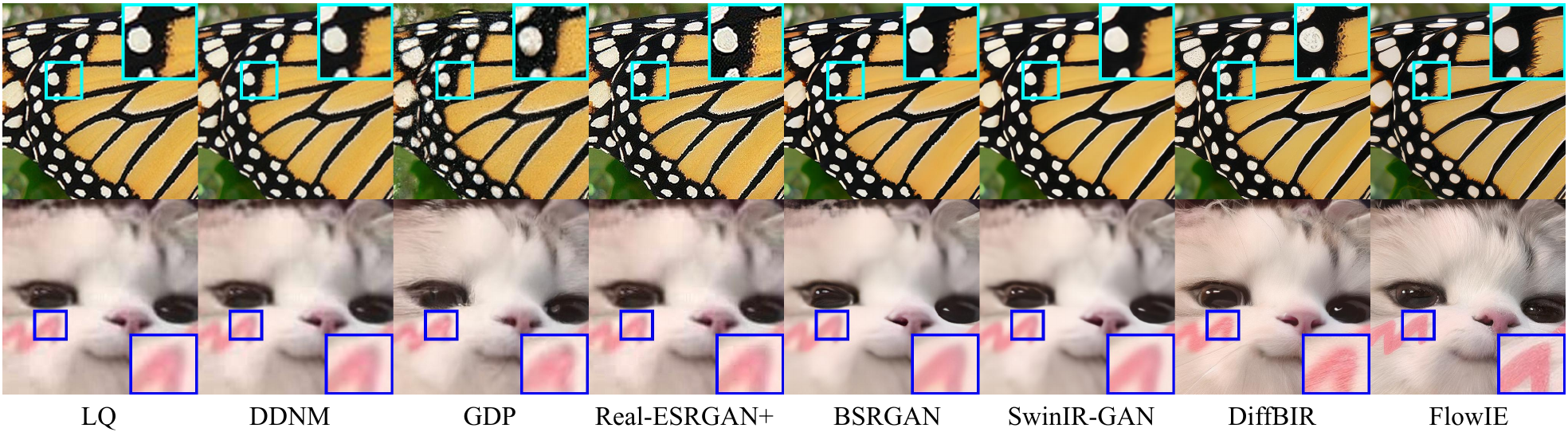}
    \vspace{-5pt}
    \caption{\textbf{Qualitative comparisons on the real-world images.} FlowIE successfully enhances the LQ images by upsampling, denoising and deblurring simultaneously and provides rich details from the generative knowledge, yielding high-quality and satisfying outcomes.}
    \label{fig:bir}
    \vspace{-5pt}
\end{figure*}

\section{Experiments}
\subsection{Experiment Setups}
\noindent \textbf{Datasets.} For the face-related tasks, including blind face restoration, face color enhancement and face inpainting, we train our model on Flickr-Faces-HQ (FFHQ)~\cite{karras2019style}, which encompasses a corpus of 70,000 high-resolution (1024 pixels) images. In preparation for training, we resize these images to a resolution of $512\times512$. To evaluate the performance of our model both quantitatively and qualitatively, we employ the synthetic CelebA-Test dataset~\cite{liu2015deep}, which comprises 3,000 pairs of LQ and HQ pairs. For comparisons on real-world datasets, we leverage LFW-Test~\cite{wang2021towards}, CelebChild-Test~\cite{wang2021towards} and WIDER-Test~\cite{zhou2022towards}, which contain face images afflicted with varying degrees of image degradations. For the blind image super-resolution task, we finetune our model on ImageNet~\cite{deng2009imagenet} and evaluate it on the widely-used RealSRSet~\cite{cai2019toward}. Since the size of RealSRSet is relatively small, we construct another test set, namely collect-100, with 100 real-world images following the class distribution in RealSRSet to conduct a broader evaluation.

\noindent \textbf{Training Details}
We apply the image restoration baseline~\cite{liang2021swinir} as our initial stage model. For BFR and BSR, we tune the initial stage model for 90K steps with a batch size of 64 on the corresponding datasets. To leverage the diffusion prior, we employ the pre-trained text-to-image model (namely "Stable Diffusion") ${\bm \epsilon}_\theta$ to initialize the path predictor ${\bm v}_\theta$ and fix the VQGAN. To optimize our rectified flow, we unfreeze the linear layers of the cross-attention blocks in ${\bm v}_\theta$ via LoRA~\cite{hu2021lora} during training. The training for these parameters takes 80K steps with a batch size of 32. For all tasks, we use the AdamW optimizer and set the learning rate as 1e-4. All tasks share the same model architecture. 

\noindent \textbf{Metrics.} To evaluate FlowIE on the blind face restoration with ground truth, we utilize traditional metrics including PSNR, SSIM and LPIPS. However these metrics are not enough to reflect human preference since they often penalize high-frequency details, \textit{e.g.}, hair texture. We also compute the identity similarity, denoted as IDS, with a face perception network~\cite{deng2019arcface} and adopt the widely-used non-reference metric FID to measure image quality, which is also employed for evaluation on wild datasets. On blind image super-resolution task, we leverage the non-reference image quality assessment metric, namely MANIQA~\cite{yang2022maniqa}, to compute image quality score with a multi-dimension attention network. To illustrate our efficiency on inference time, we calculate the throughput (FPS) of various methods.

\subsection{Main Results}
\noindent\textbf{Blind Face Restoration.} 
%Blind face restoration requires plausible mapping from LQ to the desired face image with high-quality details. 
We evaluate FlowIE on both synthetic CelebA-Test~\cite{liu2015deep} and in-the-wild LFW-Test~\cite{wang2021towards}, CelebChild-Test~\cite{wang2021towards} and WIDER-Test~\cite{zhou2022towards}.
Our comparative analysis involves recent state-of-the-art methods, including GPEN~\cite{yang2021gan}, GCFSR~\cite{he2022gcfsr}, GFPGAN~\cite{wang2021towards}, VQFR~\cite{gu2022vqfr}, RestoreFormer~\cite{wang2022restoreformer}, DMDNet~\cite{li2022learning}, CodeFormer~\cite{zhou2022towards} and DiffBIR~\cite{lin2023diffbir}. We start with the quantitative comparison on CelebA-Test, which provides LQ-HQ pairs for evaluation, as shown in Table~\ref{tab:bfr_quant}. We show that FlowIE achieves FID 19.81 and IDS 0.69, outperforming previous methods. This underscores FlowIE's effectiveness in enhancing image quality and preserving face identity. We also achieve comparable scores on PSNR, SSIM and LPIPS and exhibit a higher upper bound on all metrics than DiffBIR. Notably, the FPS of FlowIE is close to the scale of one-step methods, approximately 10 times of DiffBIR. We further showcase the qualitative results in Figure~\ref{fig:syn}. FlowIE successfully recovers detailed information like the hair and skin textures while faithfully maintaining the identity, encompassing facial features and expressions, in challenging cases. In assessing FlowIE on real-world data, we conduct experiments on three wild datasets, as presented in Table~\ref{tab:bfr_quant}. FlowIE delivers high-quality outcomes reflected by the outstanding FID on LFW-Test and WIDER-Test. We also obtain competitive FID with state-of-the-art methods on CelebChild-Test. The qualitative results on wild datasets, depicted in Figure~\ref{fig:wild}, illustrate that FlowIE consistently produces visually realistic outcomes.% with rich details, even under challenging conditions of severe blur and heavy noise encountered in real-world scenarios.
%Note that the extensive denoising steps of DiffBIR offer limited improvement for details. Instead, they introduce issues like unrealistic artifacts and color shifting, as evident in Row 1 in Figure~\ref{fig:syn} and Row 3 in Figure~\ref{fig:wild}.
\begin{table}
\centering
\caption{\textbf{Quantitative comparisons for BSR on real-world datasets.} Our flow-based framework achieves high-quality enhancement and outperforms existing methods in MANIQA with a much faster speed compared to diffusion-based methods.}
% \vspace{-10pt}
\adjustbox{width=\linewidth}{\begin{tabular}{llccc}
\toprule
\multirow{2}{*}{Type} & \multirow{2}{*}{Method}  &
\multicolumn{2}{c}{MANIQA$\uparrow$} & 
\multirow{2}{*}{FPS$\uparrow$}\\
\cmidrule{3-4}
& & RealSRSet & Collect-100 & \\
\midrule
\multirow{4}{*}{GAN} & Real-ESRGAN+~\cite{wang2021real} & 0.5373 & 0.5901 &1.875\\
& BSRGAN~\cite{zhang2021designing} & 0.5638 & 0.5889 &1.725\\
& SwinIR-GAN~\cite{liang2021swinir} & 0.5296 & 0.5721&5.978 \\
& FeMaSR~\cite{chen2022real}  &0.5250 & 0.5718 &3.167 \\
\midrule
\multirow{3}{*}{Diffusion} & DDNM~\cite{wang2022zero} & 0.4539 & 0.4813 &0.071\\
& GDP~\cite{fei2023generative} & 0.4583 & 0.5237 &0.016\\
\cmidrule{2-5}
& DiffBIR~\cite{lin2023diffbir} & 0.5906 & 0.6022 &0.286\\ 
\midrule
Flow\cellcolor{gray!25}& FlowIE (Ours)\cellcolor{gray!25}  
 & \textbf{0.5953}\cellcolor{gray!25} & \textbf{0.6087}\cellcolor{gray!25}
 &2.853\cellcolor{gray!25} \\

\bottomrule
\end{tabular}
}

\label{table:bir_quant}
\vspace{-15pt}
\end{table}

\noindent\textbf{Blind Image Super-Resolution.}
%Blind image super-resolution involves general image prior and low-level structural knowledge. 
We evaluate our FlowIE on RealSRSet~\cite{cai2019toward} and our established Collect-100 dataset. We compare FlowIE with cutting-edge methods, including GAN-based Real-ESRGAN+~\cite{wang2021real}, BSRGAN~\cite{zhang2021designing}, SwinIR-GAN~\cite{liang2021swinir}, FeMaSR~\cite{chen2022real} and diffusion-based DDNM~\cite{wang2022zero}, GDP~\cite{fei2023generative} and DiffBIR~\cite{lin2023diffbir}. In Table~\ref{table:bir_quant}, the quantitative assessment highlights FlowIE's superiority over other methods, demonstrating high image quality with MANIQA scores of 0.5953 and 0.6087 on the two datasets, respectively. Notably, though DiffBIR also attains commendable quality, its low throughput due to diffusion sampling contrasts with FlowIE's comparable speed to one-step GAN-based methods. Figure~\ref{fig:bir} demonstrates FlowIE's proficiency in enhancing intricate detailed contents, such as the patterns on the butterfly's wings in Row 1 and the texture of the cat's fur in Row 2. These vivid improvements are attributed to the generative priors from the pre-trained diffusion model. The combination of efficient inference and visually compelling results further underscores FlowIE's potential as a robust solution for challenging BSR tasks.

% It can be observed that zero-shot diffuison-based methods like DDNM~\cite{wang2022zero} and GDP~\cite{fei2023generative} fall behind in yielding clean images and GAN-based methods like Real-%ESRGAN+~\cite{wang2021real} and BSRGAN~\cite{zhang2021designing} fail to remove the nosie and generate insufficient details. 
% Though DiffBIR~\cite{lin2023diffbir} produces clean and reasonable images, it suffers from unstable noise during the sampling which may lead to artifacts like the noise on the spots of the butterfly's wings.

\subsection{Analysis}
\begin{figure}[t]

    \centering
    \includegraphics[width=0.99\linewidth]{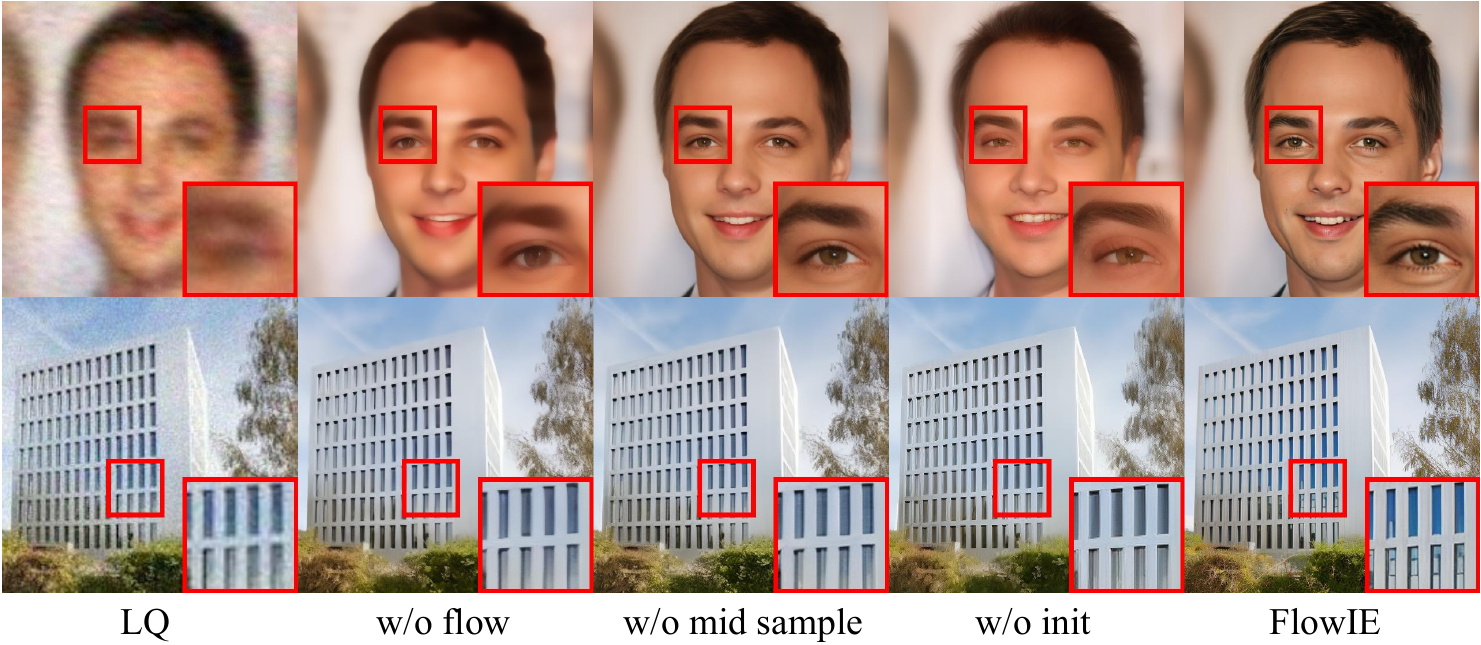}
    \setlength{\abovecaptionskip}{5pt}
    \caption{\textbf{Qualitative comparisons of the ablations.} We find the variant frameworks fall short in terms of clarity and details.}
    \label{fig:ablation}
    \vspace{-13pt}
\end{figure}

\noindent\textbf{Effective Diffusion Exploitation via Rectified Flow.} 
In diffusion models,
%the UNet ${\bm \epsilon}_\theta$ can be viewed as the learned gradient of data density
%$\nabla_{{\bm z}_t}\log p({\bm z}_t|\mathcal{C})$and 
a denoising step can be viewed as a walk along the gradient direction of data density, hinting at the potential for distilling the diffusion model to achieve faster inference. Therefore, we compare two approaches: direct distillation and rectified flow. For direct distillation (w/o flow), we set the student identical to ${\bm v}_\theta$ and fix $t=0$ during training. As shown in Table~\ref{table:ablation} and Figure~\ref{fig:ablation}, FID scores and MANIQA are adversely affected, and the visual outcomes exhibit unsatisfactory blur and inadequate details. We summarize that exploiting diffusion model via direct distillation is a tough learning problem for the one-step student model and rectified flow mitigates it with refined trajectories.  \begin{table}
\centering
\caption{\textbf{Ablation studies.} We perform ablations on BFR and BSR to verify the effectiveness of the components in FlowIE and the impact of the inference path choice. We find that rectified flow and the initial stage model are beneficial and that the path guided by mean value sampling yields the best performance.
%rectified flow, mean value sampling and the first stage model.
}
\adjustbox{width=\linewidth}
{\begin{tabular}{lcccc}
\toprule
\multirow{2}{*}{Method} & \multicolumn{2}{c}{FID$\downarrow$} & 
\multicolumn{2}{c}{MANIQA$\uparrow$} \\
\cmidrule(lr){2-3}  \cmidrule(lr){4-5} 
& CelebA &LFW & RealSRSet &Collect-100\\
\midrule
w/o flow& 49.74&53.71& 0.5311 &0.5723   \\
 w/o mid sample&25.19 &48.95  &0.5489 & 0.5805 \\
w/o init & 27.76 & 52.63& 0.5301& 0.5698\\
\midrule
\rowcolor{gray!25}
FlowIE (Ours)&\textbf{19.81} & \textbf{38.66}& \textbf{0.5953}&\textbf{0.6087} 
  
\\
\bottomrule
\end{tabular}
}

\label{table:ablation}
%\vspace{-5pt}
\end{table}

\noindent\textbf{Choice of Inference Paths.}
The straightened path via rectified flow is not perfectly linear. Empirically, we have two options for inference paths: (1) use the forward Euler method which walks along the trajectory with fixed step length (w/o mid sample), and (2) follow Lagrange's Mean Value Theorem to identify a pivotal midpoint on the path. We generate the results for BFR and BSR through both paths. Results from Path 1, as depicted in Table~\ref{table:ablation} and Figure~\ref{fig:ablation}, show plausible outcomes with reduced noise. However, they fall behind in realness and details compared to Path 2. We conclude that the Euler method struggles to produce high-quality images in very few steps (\textit{e.g.}, 5), while our mean value sample obtains visually pleasant results with a more efficient inference process (\textless5 steps).

% \noindent\textbf{Unfreeze Strategy for Cross-Attention Layers.} 

\noindent\textbf{Impact of the Initial Stage Model.} Our initial stage model ${\bm \tau}_\phi$ performs general image deblurring on LQ images, enhancing guidance quality and thereby strengthening FlowIE's path estimation. To gauge this impact, we train our rectified flow without ${\bm \tau}_\phi$ (w/o init) and conduct an evaluation on test sets. We observe that the absence of ${\bm \tau}_\phi$ results in low-quality guidance, leading to unsatisfactory results. This is evident in the worse FID and MANIQA scores in Table~\ref{table:ablation} and blurred object edges in Figure~\ref{fig:ablation}. These results demonstrate the effect of ${\bm \tau}_\phi$ in improving the quality of conditions and achieving better overall performance.   
\subsection{Extensions}
To further demonstrate the adaptability of our framework, we generalize FlowIE to extended tasks, including face color enhancement and face inpainting. Achieving this extension requires a minimal fine-tuning effort of 5K steps for the rectified flow dedicated to each task.%We will describe our implementation and provide qualitative results. Additionally, more results on other enhancement tasks and datasets can be found in the supplementary material.  

\begin{figure}[t]
    \centering
    \includegraphics[width=0.99\linewidth]{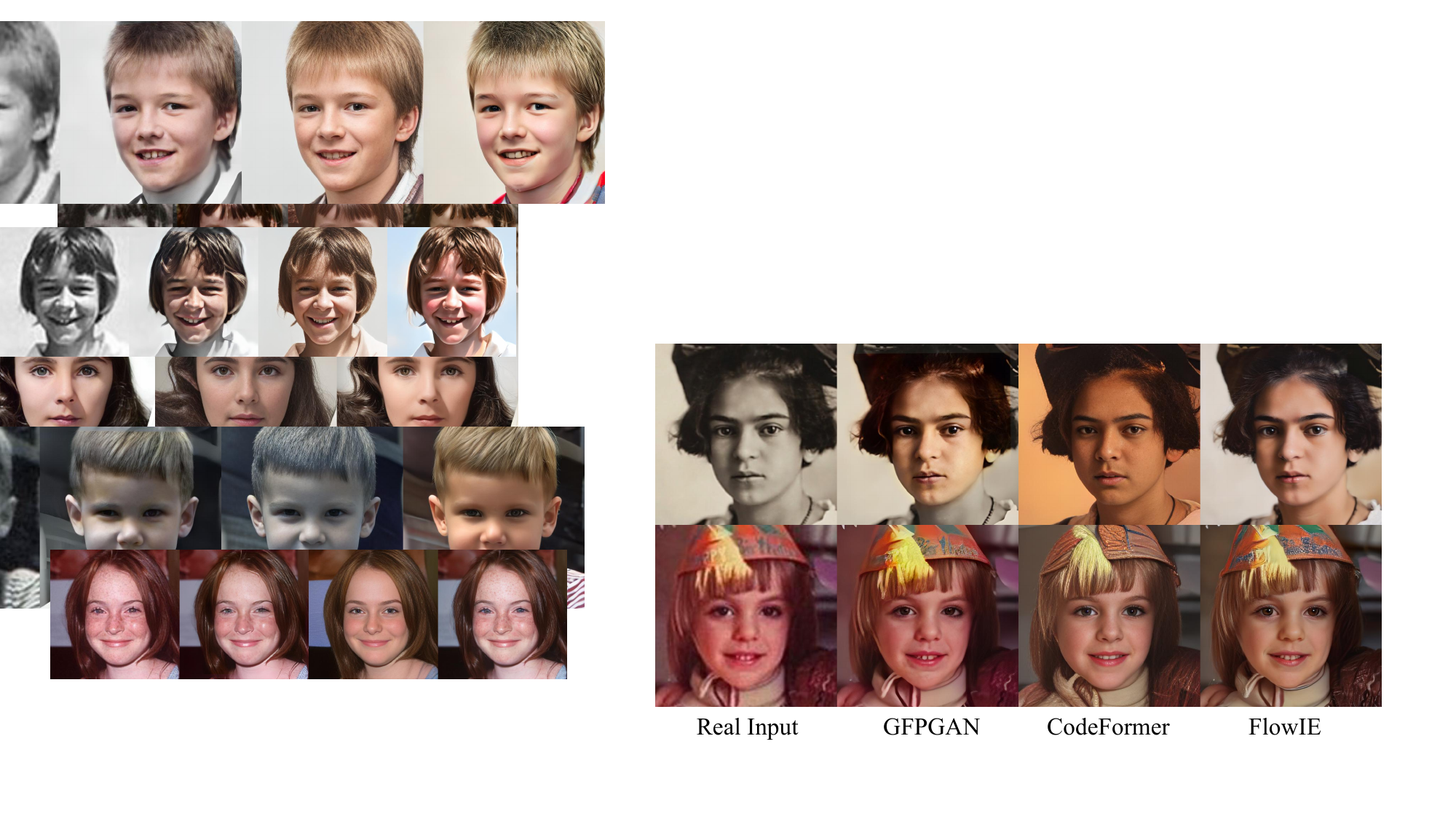}
    \setlength{\abovecaptionskip}{5pt}
    \caption{\textbf{Face color enhancement via \textit{FlowIE}}. We yield satisfying enhancement results with vivid colors for the old photos.}
    \label{fig:color}
    \vspace{-5 pt}
\end{figure}

\noindent\textbf{Face Color Enhancement.} %Due to age and device limitations, old photos often experience color shifts or only have the grayscale channel. Color Enhancement aims to recover these old photos with natural colors. 
To achieve color enhancement, we fix the initial stage model and fine-tune our rectified flow using color augmentations (random color jitter and grayscale conversion) in~\cite{wang2021towards}. We compare our method with GFPGAN~\cite{wang2021towards} and CodeFormer~\cite{zhou2022towards} on real-world CelebChild-Test~\cite{wang2021towards} dataset. The results in Figure~\ref{fig:color} showcase that FlowIE produces visually appealing and highly consistent face images with vibrant and realistic colors.
%, while other methods fail to colorize the grayscale image and generate unexpected artifacts and inconsistency in face shape and expression.    
\begin{figure}[t]
    \centering
    \includegraphics[width=0.99\linewidth]{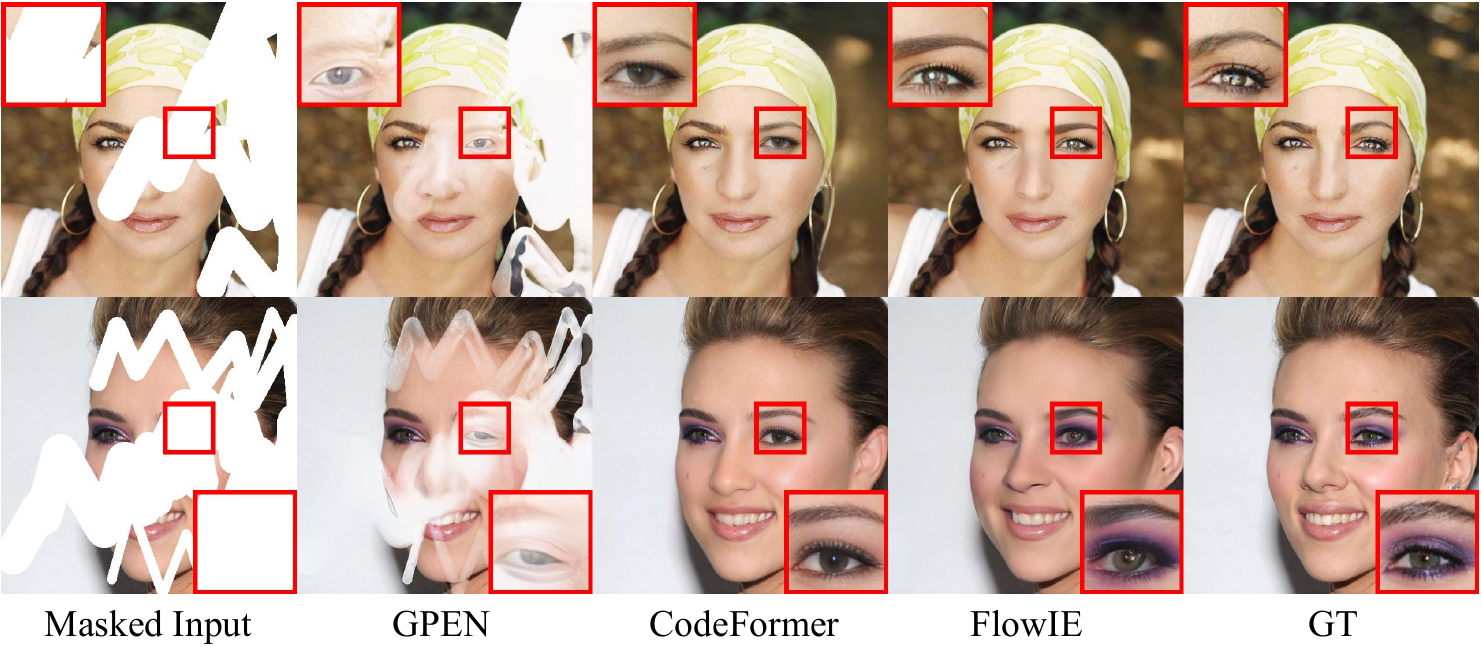}
    \setlength{\abovecaptionskip}{5pt}
    \caption{\textbf{Face inpainting via \textit{FlowIE}}. We complete the missing pixels with realistic and coherent content for 
 challenging cases.}
    \label{fig:inp}
    \vspace{-15 pt}

\end{figure}

\noindent\textbf{Face Inpainting.} We employ the script from~\cite{yang2021gan} to draw irregular polyline masks on face images as our inputs and fine-tune the rectified flow. During inference, we resize the mask to the latent code's shape and use it to maintain the visible area on the inputs. As shown in Figure~\ref{fig:inp}, FlowIE successfully reconstructs the challenging cases and seamlessly completes them with coherent contents.%, illustrating its profound understanding of semantic-level facial parts. %In comparison, GPEN~\cite{yang2021gan} fails to entirely remove the mask and CodeFormer~\cite{wang2021towards} generates visible artifacts lying in unnatural eyes and asymmetrical earrings.

\section{Conclusion}
In this paper, we introduced FlowIE, a novel framework that harnesses the conditioned rectified flow to exploit the potent generative priors within the pre-trained diffusion model and accelerate the inference by straightening the probability transport trajectories. To further improve the path estimation accuracy and reduce inference steps, we have devised the mean value sampling to predict a precise direction at the curve midpoint. %Extensive experiments showcased that our framework achieves very competitive performance and exhibits remarkable generalization capabilities, adapting seamlessly to various image enhancement challenges.
Extensive experiments demonstrate our framework's competitive performance and remarkable generalization across diverse image enhancement challenges. We envision our work will inspire future research on flow-based image enhancement and efficient diffusion sampling.%, driving progress in real-world applications.
%ultimately leading to advancements in real-world applications.

\noindent
\textbf{Acknowledgement.}
This work was supported in part by the National Natural Science Foundation of China under Grant 62125603, Grant 62321005, and Grant 62336004.
{
    \small
    \bibliographystyle{ieeenat_fullname}
    \bibliography{main}
}
%\appendix % Start the appendix section
\renewcommand\thesection{\Alph{section}} 
\renewcommand\thetable{\Alph{table}}
\renewcommand\thefigure{\Alph{figure}}

\setcounter{section}{0}
\setcounter{figure}{0}
\setcounter{table}{0}

\clearpage
\setcounter{page}{1}
\section*{Appendix}
In the supplementary material, we provide a deeper exploration of insights and findings. In Section~\ref{sec:a}, we present more implementation details regarding the training and evaluation of FlowIE. Section~\ref{sec:b} delves into further discussions through a combination of quantitative analyses and qualitative experiments. In Section~\ref{sec:c}, we show additional visualization results for Blind Face Restoration (BFR) and Blind Image Super-Resolution (BSR). Furthermore, Section~\ref{sec:d} extends our investigations to encompass tasks such as single image deraining and dehazing. The source code for FlowIE is also provided in the zip file.

%  A
\section{Detailed Implementations}
\label{sec:a}
To initialize our path estimator ${\bm v}_\theta$, we employ the text-to-image pretrained Stable Diffusion (SD 2.0-Base)~\cite{rombach2022high}, which offers ample generative priors suitable for various enhancement tasks. The input image ${\bm x}\in\mathbb{R}^{3\times512\times512}$ is encoded into the latent code ${\bm z}\in\mathbb{R}^{4\times64\times64}$ by the trained VQGAN. During the training of all tasks, we resize the input images to $512\times512$. For the images smaller than this size, we upsample them with the short side enlarged to $512$ and crop them with a fixed-size bounding box. We train our FlowIE with 8 NVIDIA RTX 3090 GPUs. To maintain the pre-trained capability of the diffusion model, we utilize the LoRA~\cite{hu2021lora} approach to unfreeze the linear layers of the cross-attention blocks in ${\bm v}_\theta$. We find that the small trainable parameters with a LoRA rank of 4 can significantly unleash the generative priors within the diffusion model and allow adaptation to various tasks. Another benefit of the partially unlocked models is preventing overfitting of the large diffusion model. To measure the throughput of FlowIE and other methods, we conduct evaluation experiments on the same dataset and using a single 3090 GPU.

\section{More Discussions}
\label{sec:b}
\begin{figure}[t]
    %\vspace{-5pt}
    \centering
    \includegraphics[width=0.99\linewidth]{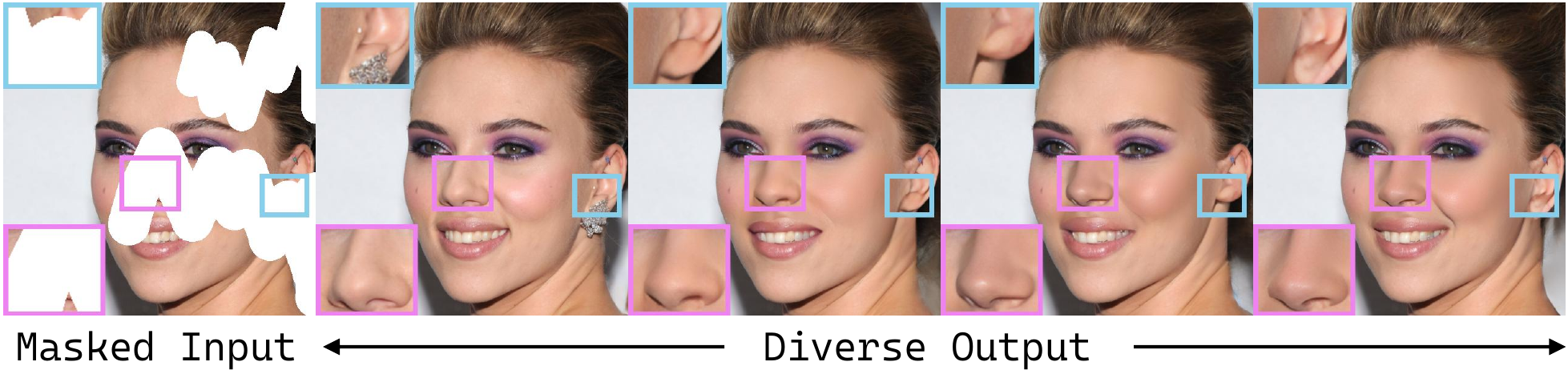}
    \setlength{\abovecaptionskip}{5pt}
    \caption{\textbf{Diverse results of \textit{FlowIE}}. Our framework can generate various results with different initial noises.}
    \label{fig:diverse}
    \vspace{-5pt}
\end{figure}

\noindent\textbf{Many-to-one mapping and result diversity.} Compared with text-to-image generation, image enhancement tasks like BFR have more deterministic targets. Therefore, we employ the `many-to-one' strategy for FlowIE during training to learn a direct mapping from noise to real data. However, it's crucial to clarify that FlowIE, being a probabilistic model like diffusion models, inherently yields diverse outcomes, especially for the inpainting task. As illustrated in Figure~\ref{fig:diverse}, given the masked input (Col.1) and different initial noise ${\bm z}_0$, FlowIE generates various facial features (Col.2-5), encompassing variations in the shape of the nose, ears, and texture of the hair. Unlike rigid `many-to-one' mapping often employed in GAN-based methods during inference, FlowIE embraces the generative capacity of diffusion models and enjoys the diversity of plausible results.   

\begin{figure}[t]
    \centering
    \includegraphics[width=0.99\linewidth]{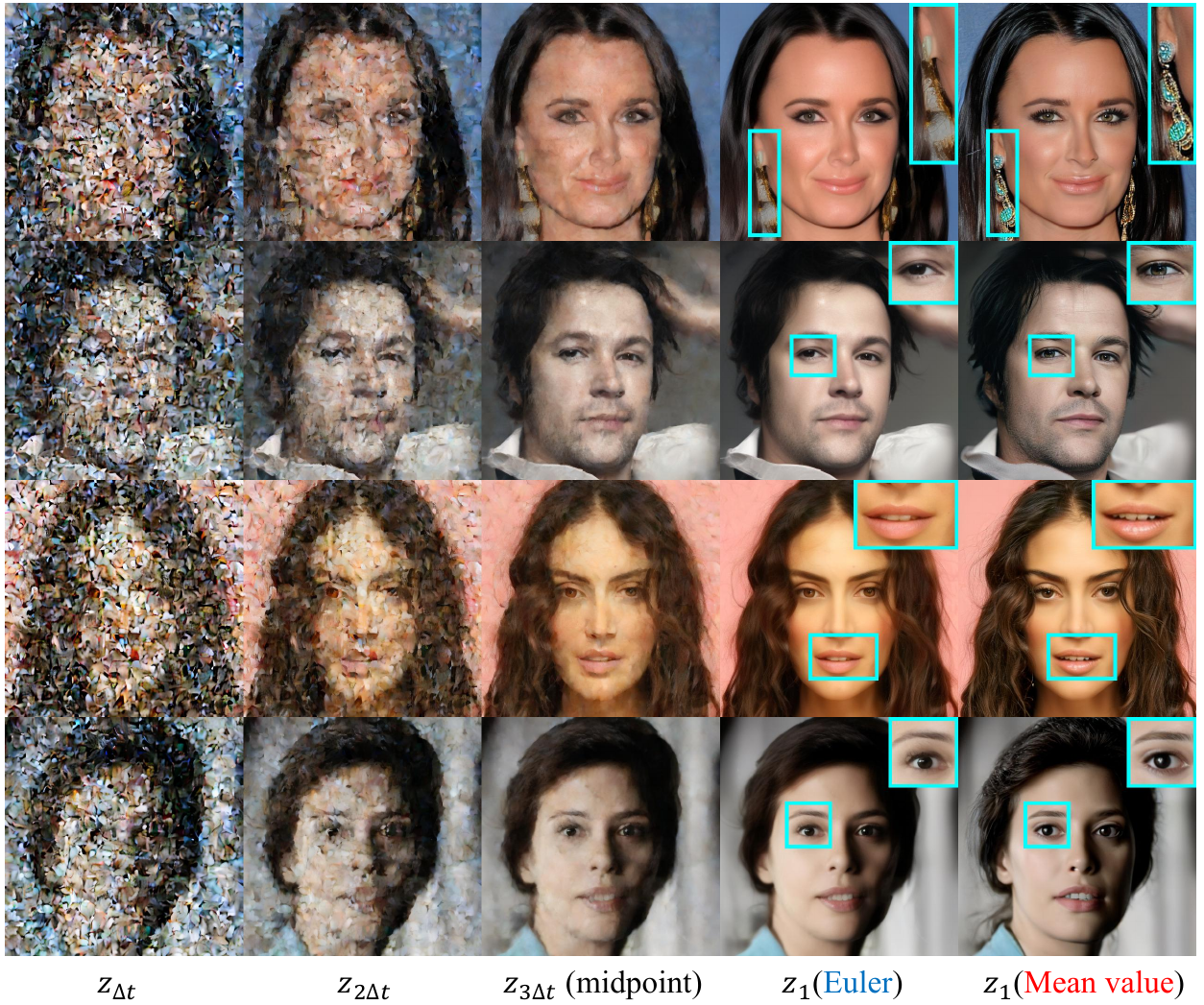}
    \caption{\textbf{The visualization of the inference process.} FlowIE establishes straight-line paths from random noise to clean images. Through mean value sampling, we achieve clearer and more detailed results in fewer steps compared to the Euler method.}
    \label{fig:supp_paths}
    \vspace{-5pt}
\end{figure}
\begin{figure*}[t]
    \centering
    \includegraphics[width=0.99\linewidth]{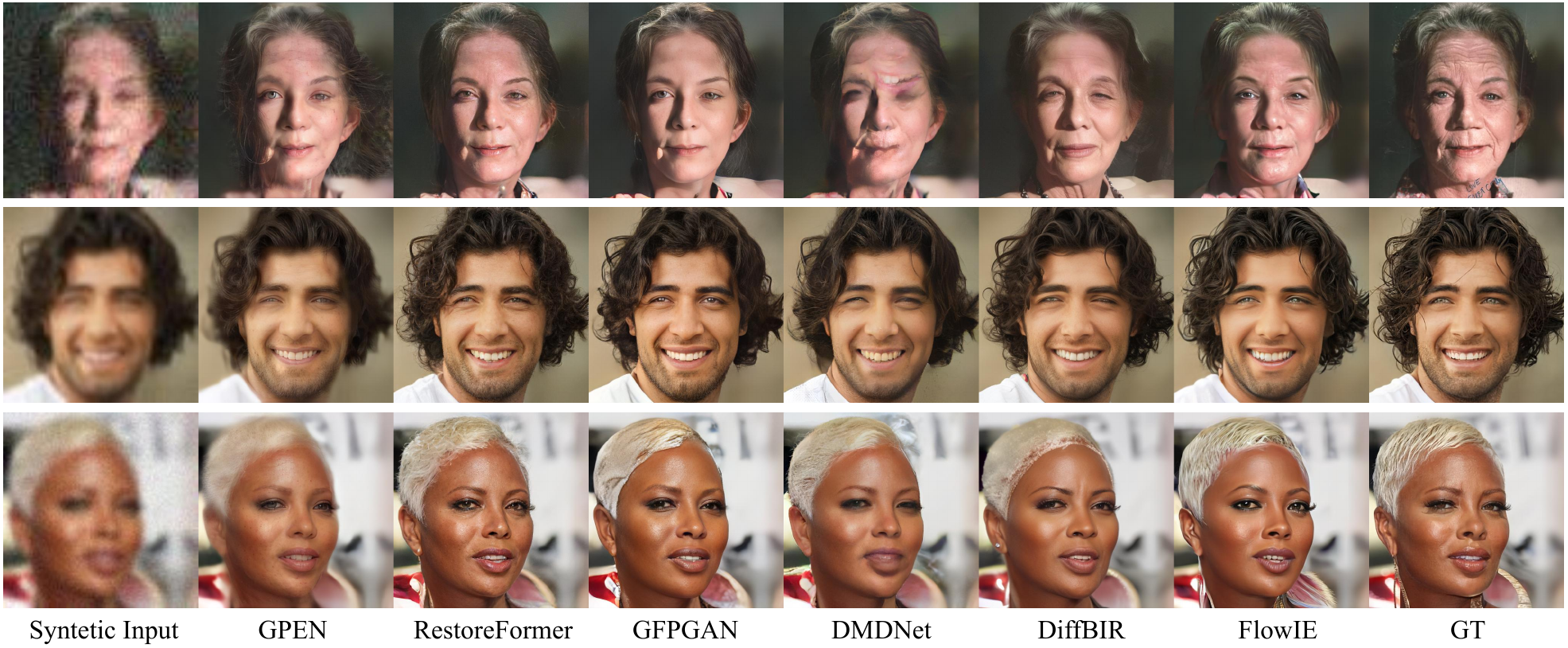}
    \caption{\textbf{Qualitative comparisons on CelebA-Test.} FlowIE produces high-quality results with rich details and maintains high identity similarity, even when confronted with severely degraded inputs, while previous methods exhibit visible artifacts or inconsistent faces.}
    \label{fig:syn_more}
    
\end{figure*}
\begin{figure*}[t]
    \centering
    \includegraphics[width=0.99\linewidth]{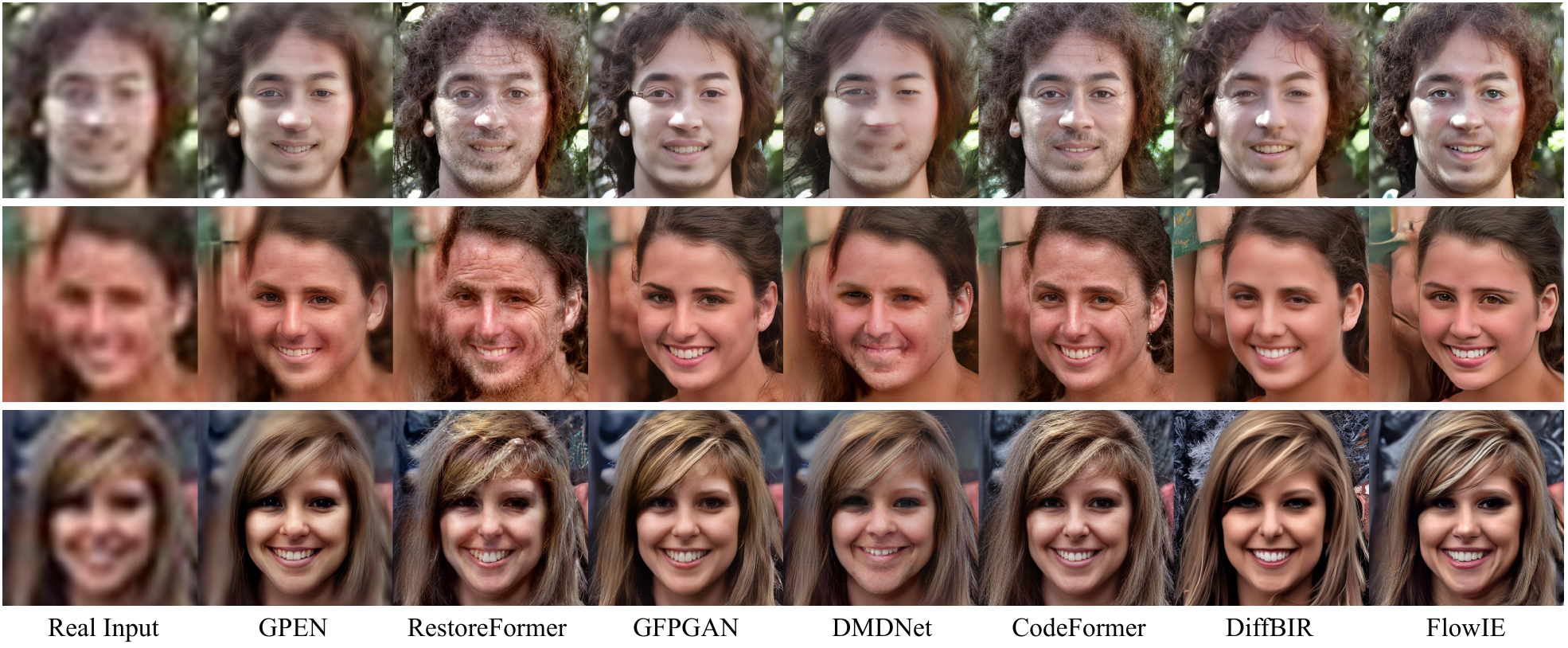}
    \caption{\textbf{Qualitative comparisons on real-world faces.} Our approach demonstrates credible enhancements on real-world faces, delivering high-fidelity and visually satisfying results. Compared to other methods, FlowIE showcases robustness in front of challenging cases.}
    \label{fig:wild_more}
    \vspace{-10pt}
\end{figure*}
%% B

\noindent\textbf{Visualization of different paths} We showcase the visualization of each step in our inference process. Along the straight-line path, FlowIE adeptly generates high-quality (HQ) images from noise in less than 5 steps. As depicted in Figure~\ref{fig:supp_paths}, the mean value sampling consistently yields clearer and more detailed results in fewer steps compared to the Euler method, highlighting its efficacy in enhancing the quality of the generated images.

\noindent\textbf{About starting from ${\bm\tau}_{\phi}({\bm z}_{\rm LQ})$.} Since FlowIE predicts the path from random noise, switching the starting point to the coarse result ${\bm\tau}_{\phi}({\bm z}_{\rm LQ})$ is indeed reasonable. Tuning and evaluation on the BFR task (shown in Table~\ref{table:start}) indicate a slightly worse FID compared with FlowIE. We attribute this result to the adjustment's reliance on initial results over pre-trained diffusion priors.
\begin{table}[!h]
%\begin{wraptable}{r}{4.3cm}
\centering
\caption{\textbf{Ablation study about the starting point.} Latent initiation from ${\bm z}_0={\bm\tau}_{\phi}({\bm z}_{\rm LQ})$ leads to worse FID.}
%\vspace{-10pt}
%\adjustbox{width=\linewidth}
%{
\scriptsize
%\hspace{-.3in}
\begin{tabular}{lcc}
\toprule
\multirow{2}{*}{Method} & \multicolumn{2}{c}{FID$\downarrow$} \\
\cmidrule(lr){2-3}   
& CelebA &LFW\\
% \midrule
% ${\bm\tau}_{\phi}({\bm z}_{\rm LQ})$ (2 steps)& \textbf{21.37}&\textbf{41.81} \\
% \rowcolor{gray}
% noise (2 steps) & 27.76 & 52.63\\
\midrule
DiffBIR~\cite{lin2023diffbir} & 20.19&39.61 \\
${\bm z}_0={\bm\tau}_{\phi}({\bm z}_{\rm LQ})$ & 19.87 &38.80 \\
\rowcolor{gray}
FlowIE &\textbf{19.81} & \textbf{38.66}
  
\\
\bottomrule
\end{tabular}
%}

\label{table:start}
\vspace{-10pt}
\end{table}
%\end{wraptable}

\begin{figure}[t]
    
    \centering
    \includegraphics[width=0.99\linewidth]{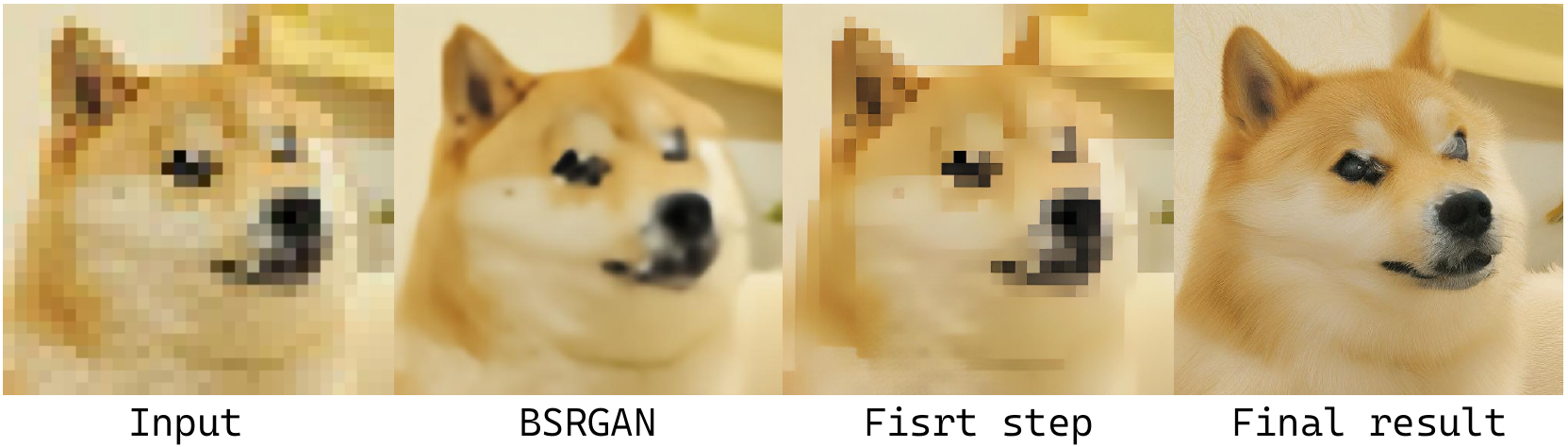}
    \setlength{\abovecaptionskip}{5pt}
    \caption{\textbf{Failure case of \textit{FlowIE}}. Our framework may give unsatisfying results when facing severe degradation.}
    \label{fig:artifact}
    \vspace{-10pt}
\end{figure}

\noindent\textbf{About artifacts in the first step.} We acknowledge that extreme artifacts in the first step may result in failure cases. In Figure~\ref{fig:artifact}, the input undergoes challenging degradation ($16\times$ downsampling). Compared to GAN-based methods like BSRGAN~\cite{zhang2021designing} which introduce many artifacts and blur, FlowIE generates a cleaner image. However, the final result may still exhibit unrealistic eyes due to initial step artifacts. 

\begin{figure}[t]
    % \vspace{-10pt}
    \centering
    \includegraphics[width=0.99\linewidth]{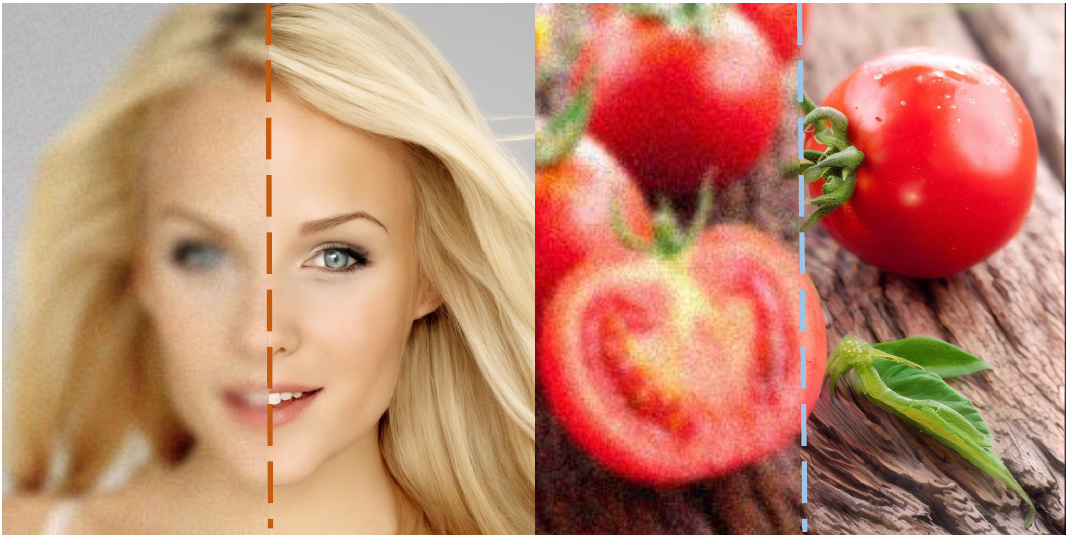}
    \setlength{\abovecaptionskip}{5pt}
    \caption{\textbf{Qualitative results in larger resolution.} The proposed FlowIE consistently delivers visually captivating results at higher resolutions.}
    \label{fig:1024}
    \vspace{-10pt}
\end{figure}

\noindent\textbf{About larger resolution.} FlowIE demonstrates excellent scalability to process larger images. We can replace the original diffusion model (SD 2.0-base) with an enlarged version (SDXL) which generates $1024\times1024$ images by default and tune the FlowIE framework following the proposed method. As shown in Figure~\ref{fig:1024}, despite the limitation in training time, we still obtain satisfying outcomes with higher resolution ($1024\times1024$). %We also conduct a patch-based sampling strategy to handle even larger images.    

\noindent\textbf{Comparisons with diffusion models.} Our proposed FlowIE mainly capitalizes on the powerful generation capability within the pre-trained diffusion model, which has demonstrated its versatility in various visual tasks. For example, DDVM~\cite{saxena2024ddvm} explicitly underscores the effectiveness of pre-trained priors in diffusion models for monocular depth estimation and SDEidt~\cite{meng2021sdedit} focuses on image editing tasks like stroke-based editing. Additionally,~\cite{xiao2021tackling} successfully achieves rapid image sampling by employing
multimodal denoising distributions and conditional GANs. Compared with these works, our FlowIE primarily harnesses the generative prior in diffusion models and employs a conditioned flow-based strategy to accelerate the sampling.
% C
\begin{figure}[!h]
    \centering
    \includegraphics[width=\linewidth]{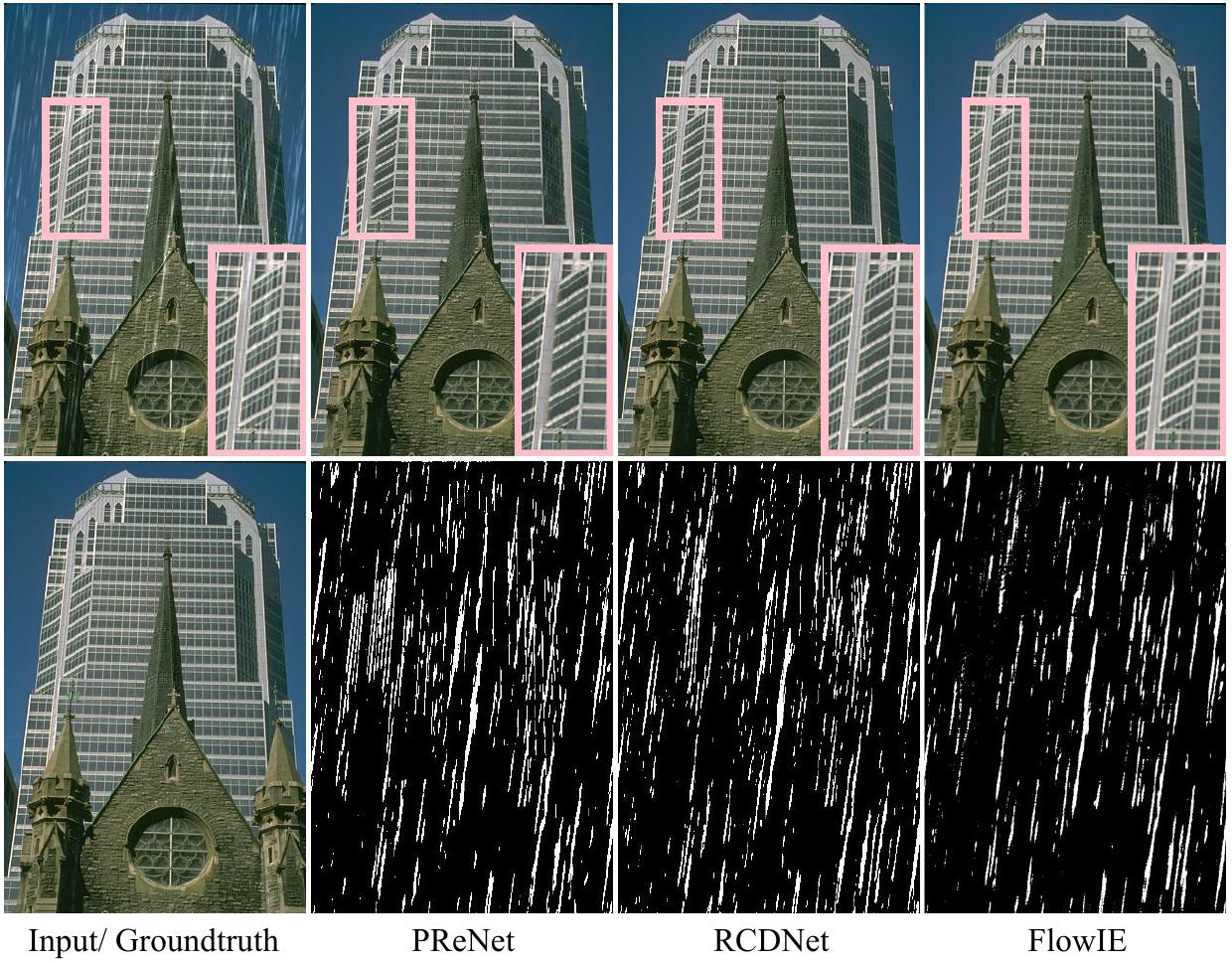}
    \caption{\textbf{Single image Deraining via \textit{FlowIE}.} Our framework adeptly identifies the rainy layers and proficiently restores the original images without complex task-specific designs. 
    }
    \label{fig:derain}
    
\end{figure}
\begin{figure}[!h]
    \centering
    \includegraphics[width=\linewidth]{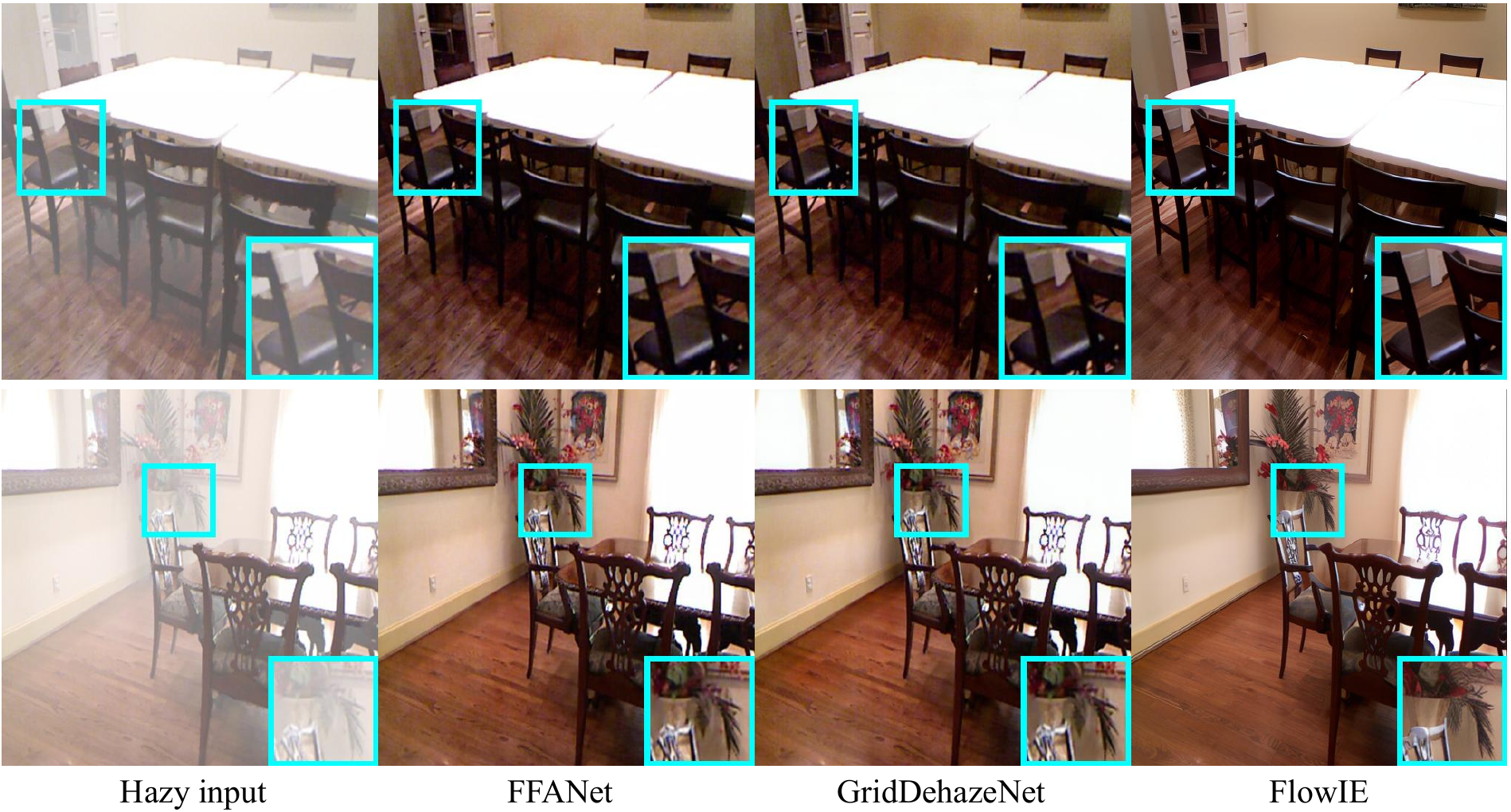}
    \caption{\textbf{Single image dehazing via \textit{FlowIE}.} Our framework effectively eliminates haze, enhancing the overall clarity and quality of the images.}
    \vspace{-10pt}
    \label{fig:dehaze}
    
\end{figure}
\section{More Qualitative Comparisons}
\label{sec:c}

In this section, we provide additional visual comparisons on BFR and BSR with state-of-the-art methods. Our framework reliably demonstrates its ability to deliver robust and satisfying results in these challenging tasks, showcasing its efficacy across diverse image enhancement scenarios.
\begin{figure*}[t]
    \centering
    \includegraphics[width=0.99\linewidth]{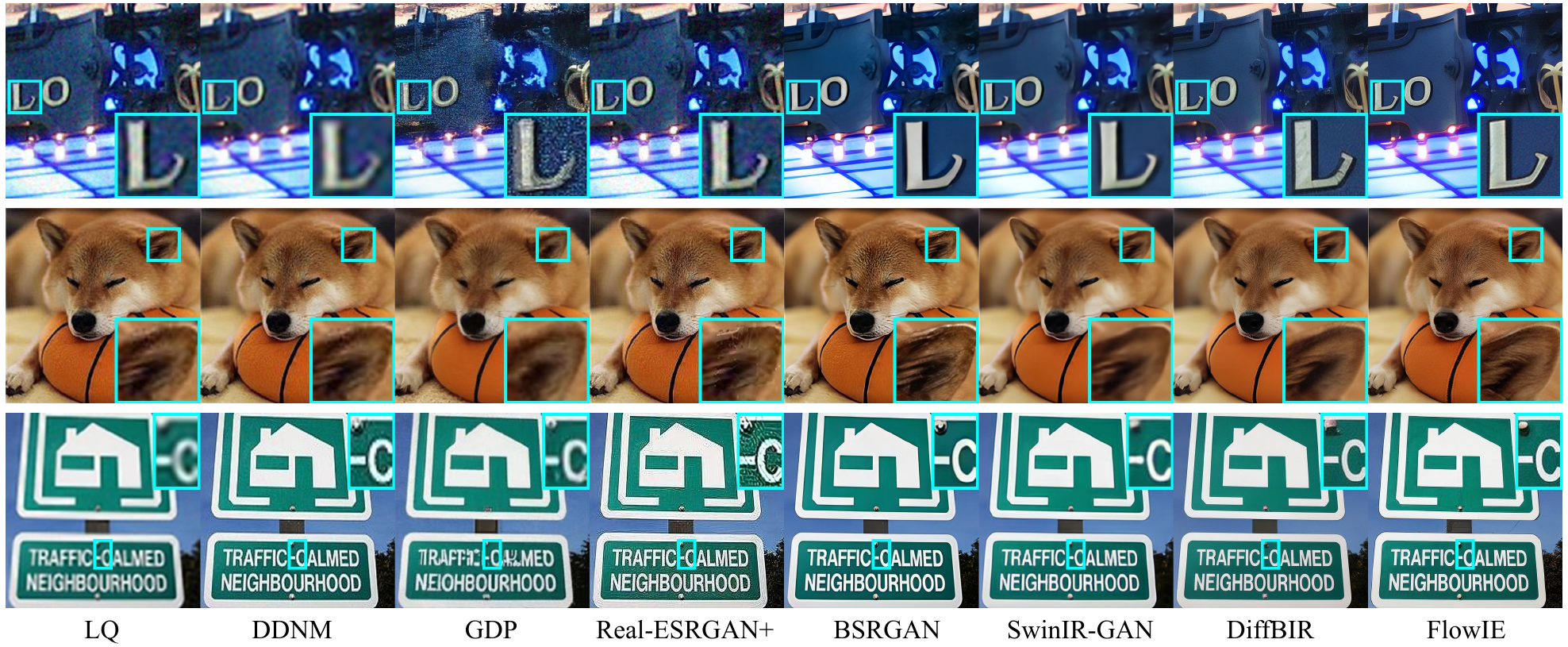}
    \caption{\textbf{Qualitative comparisons on the real-world images.} FlowIE successfully enhances the LQ images through simultaneous upsampling, denoising, and deblurring, and provides rich details from the generative knowledge, leveraging generative knowledge to deliver high-quality outcomes with rich details.}
    \label{fig:bir_more}
    
\end{figure*}

\noindent\textbf{Blind Face Restoration.}
We conduct qualitative comparisons on both synthetic CelebA-Test~\cite{liu2015deep} and in-the-wild LFW-Test~\cite{wang2021towards}, CelebChild-Test~\cite{wang2021towards} and WIDER-Test~\cite{zhou2022towards}.
Our comparisons involve recent state-of-the-art methods, including GPEN~\cite{yang2021gan}, GCFSR~\cite{he2022gcfsr}, GFPGAN~\cite{wang2021towards}, VQFR~\cite{gu2022vqfr}, RestoreFormer~\cite{wang2022restoreformer}, DMDNet~\cite{li2022learning}, CodeFormer~\cite{zhou2022towards} and DiffBIR~\cite{lin2023diffbir}. Visual results presented in Figure~\ref{fig:syn_more} and Figure~\ref{fig:wild_more} demonstrate that our FlowIE consistently produces visually pleasing outcomes on both synthetic and real-world datasets, affirming its effectiveness and robust performance in diverse scenarios.

\noindent\textbf{Blind Image Super-Resolution.}
For BSR, we also present additional results on RealSRSet~\cite{cai2019toward} and our established Collect-100 dataset. We compare FlowIE with cutting-edge methods, including GAN-based Real-ESRGAN+~\cite{wang2021real}, BSRGAN~\cite{zhang2021designing}, SwinIR-GAN~\cite{liang2021swinir}, FeMaSR~\cite{chen2022real} and diffusion-based DDNM~\cite{wang2022zero}, GDP~\cite{fei2023generative} and DiffBIR~\cite{lin2023diffbir}. Figure~\ref{fig:bir_more} vividly illustrates the efficacy of FlowIE in generating visually appealing images with a commendable balance between realism and clarity.

% D
\section{More Extended Tasks}
\label{sec:d}
To showcase the versatility of our framework, we extend FlowIE to more tasks, specifically single image deraining and dehazing. The adaptation for these tasks involves a fine-tuning process with 15K steps on the respective datasets. Notably, we only use the single MSE loss for all tasks.

\noindent\textbf{Deraining.}
We utilize RainTrainH~\cite{rain-trainH}, RainTrainL~\cite{rain-trainH} and Rain12600~\cite{rain12600} for training and evaluate our framework on Rain-100L dataset~\cite{rain100l}. We compare our results with PReNet~\cite{PReNet} and RCDNet~\cite{rcdnet}. As shown in Figure~\ref{fig:derain}, FlowIE effectively separates the rainy layers and reconstructs the original clean images.

\noindent\textbf{Dehazing.} We employ the indoor part of the RESIDE dataset~\cite{RESIDE} for training and evaluate our framework on its test split. We compare the results with FFA-Net~\cite{ffa-net} and GridDehazeNet~\cite{griddehaze}. FlowIE demonstrates successful haze removal and enhances the clarity of the original images, as shown in Figure~\ref{fig:dehaze}.

%\\[80pt]

% \section{Rationale}
% \label{sec:rationale}
% % 
% Having the supplementary compiled together with the main paper means that:
% % 
% \begin{itemize}
% \item The supplementary can back-reference sections of the main paper, for example, we can refer to \cref{sec:intro};
% \item The main paper can forward reference sub-sections within the supplementary explicitly (e.g. referring to a particular experiment); 
% \item When submitted to arXiv, the supplementary will already included at the end of the paper.
% \end{itemize}
% % 
% To split the supplementary pages from the main paper, you can use \href{https://support.apple.com/en-ca/guide/preview/prvw11793/mac#:~:text=Delete%20a%20page%20from%20a,or%20choose%20Edit%20%3E%20Delete).}{Preview (on macOS)}, \href{https://www.adobe.com/acrobat/how-to/delete-pages-from-pdf.html#:~:text=Choose%20%E2%80%9CTools%E2%80%9D%20%3E%20%E2%80%9COrganize,or%20pages%20from%20the%20file.}{Adobe Acrobat} (on all OSs), as well as \href{https://superuser.com/questions/517986/is-it-possible-to-delete-some-pages-of-a-pdf-document}{command line tools}.
% \clearpage


\end{document}